\documentclass{article}
\usepackage{ijcai25}

\usepackage{times}
\usepackage{soul}
\usepackage{url}
\usepackage[hidelinks]{hyperref}
\usepackage[utf8]{inputenc}
\usepackage[small]{caption}
\usepackage{graphicx}
\usepackage{amsmath}
\usepackage{amsthm}
\usepackage{booktabs}
\usepackage{algorithm}
\usepackage{algorithmic}
\usepackage[switch]{lineno}
\usepackage{xcolor}
\usepackage{multirow}
\usepackage{subcaption}
\usepackage{amssymb}

\usepackage{color}

\title{
Exploring Visual Prompting: Robustness Inheritance and Beyond
}

\author{
{\rm Qi Li$^1$, 
Liangzhi Li\thanks{Corresponding author.}$^{2}$
Zhouqiang Jiang$^2$, 
Bowen Wang$^2$,
Keke Tang$^3$\\
$^1$National University of Singapore\\
$^2$Osaka University \\
$^3$The University of Hong Kong\\ 
liqi@u.nus.edu,
li@ids.osaka-u.ac.jp
}
}

\begin{document}

\maketitle

\begin{abstract}
Visual Prompting (VP), an efficient method for transfer learning, has shown its potential in vision tasks. However, previous works focus exclusively on VP from standard source models, it is still unknown how it performs under the scenario of a robust source model: Can the robustness of the source model be successfully inherited? Does VP also encounter the same trade-off between robustness and generalization ability as the source model during this process? If such a trade-off exists, is there a strategy specifically tailored to VP to mitigate this limitation? In this paper, we thoroughly explore these three questions for the first time and provide affirmative answers to them. To mitigate the trade-off faced by VP, we propose a strategy called Prompt Boundary Loosening (PBL). As a lightweight, plug-and-play strategy naturally compatible with VP, PBL effectively ensures the successful inheritance of robustness when the source model is a robust model, while significantly enhancing VP's generalization ability across various downstream datasets. Extensive experiments across various datasets show that our findings are universal and demonstrate the significant benefits of the proposed strategy.
\end{abstract}

\section{Introduction}
\label{sec:intro}

Transferring knowledge from large-scale datasets enables efficient learning for new tasks~\cite{pan2009survey,chen2021exploring,bao2021beit}, among which various paradigms that leveraging pre-trained models, such as fine-tuning~\cite{howard2018universal,kumar2022fine} and linear probing have been widely adopted. While effective, these methods typically require parameter tuning or architectural modifications, leading to high computational costs and limited generalizability.

To address these challenges, Visual Prompting (VP)~\cite{bahng2022exploring} or model reprogramming~\cite{tsai2020transfer,elsayed2018adversarial} has emerged as a lightweight and efficient alternative for knowledge transfer. VP keeps the pre-trained model frozen and instead learns a small set of parameters as input prompts. This approach not only reduces computational overhead but also facilitates adaptability across diverse tasks without altering the underlying model.

\begin{figure}[t!]
  \centering
   \includegraphics[width=\columnwidth]{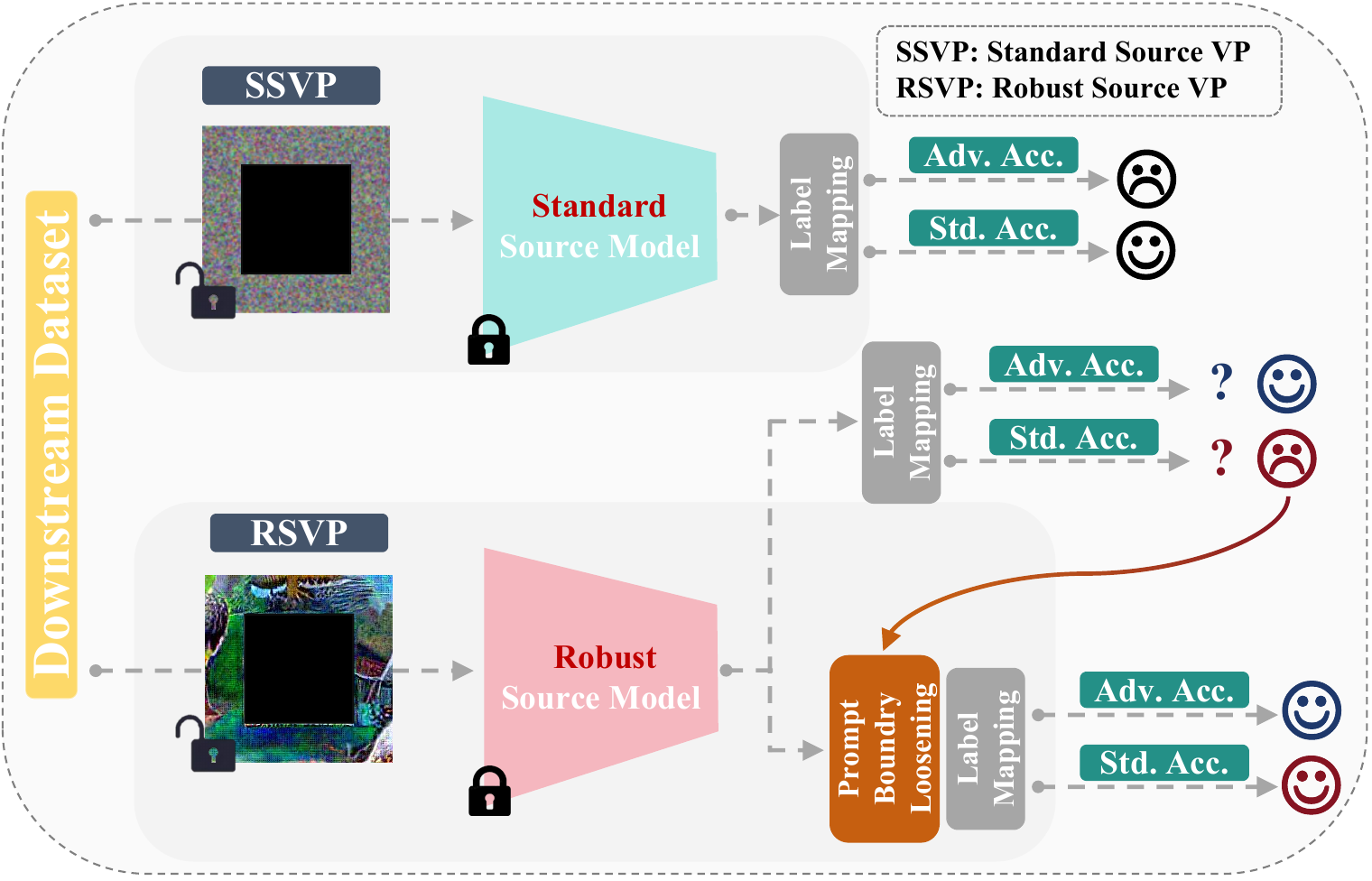}
   \caption{RSVP can inherit the robustness from the source model while also suffer from generalization degradation. RSVP are visually more human aligned. The proposed PBL brings RSVP a better trade-off between robustness and generalization.}
   \label{fig:teaser}
\end{figure}

However, as shown in Figure \ref{fig:teaser}, existing research predominantly uses standard-trained models obtained without adversarial training, which are highly susceptible to adversarial attacks~\cite{goodfellow2014explaining,chakraborty2018adversarial}. On the other hand, robust models trained with adversarial training~\cite{shafahi2019adversarial,ganin2016domain} offer resilience against such attacks but often suffer from degraded standard accuracy~\cite{tsipras2018robustness,gowal2020uncovering}. Furthermore, the process of adversarial training is computationally expensive due to its bi-level optimization process~\cite{wong2020fast,wang2019bilateral}. Considering the good generalization ability of the Standard Source VP (SSVP) and its lightweight in training, it is meaningful to study the properties of Robust Source VP (RSVP). 

In this work, we explore RSVP as a promising yet underexplored scenario. Specifically, we aim to address three fundamental questions: i) Can RSVP inherit the robustness of its robust source model? ii) Does RSVP also experience suboptimal generalization performance similar to its source model? iii) How can we explain these phenomena and mitigate potential limitations?

Our findings reveal that RSVP inherits both the robustness of the source model and its generalization challenges. To explain this, we analyze RSVP's visual representations, showing that it visually aligns better with human perception. To address the negative transfer effect of RSVP on generalization performance, we propose a plug-and-play strategy named Prompt Boundary Loosening (PBL), which extends the mapping range of each label in downstream tasks while preserving the complex decision boundaries of robust models. This strategy not only maintains robustness but also significantly enhances generalization performance.

Overall, our contribution is summarized as follows:
\begin{itemize}
    \item We pioneer the exploration of Robust Source VP (RSVP), identifying its strengths in inheriting robustness and its limitations in generalization performance.
    \item We provide a comprehensive explanation of RSVP's behavior through an analysis of visual representations. We find that RSVP are visually more human-aligned and usually contains some texture patterns, bridging the gap between understanding the behavior of RSVP and adversarial training.
    \item We propose Prompt Boundary Loosening (PBL), a novel strategy that improves RSVP’s generalization without compromising (and often enhancing) its robustness. Extensive experiments demonstrate the universality of RSVP’s characteristics and the effectiveness of PBL across diverse datasets.
\end{itemize}


\section{Related Work}
\label{sec:RelatedWork}
{\bf Prompt Learning in vision tasks.}
Given the success of prompt tuning in natural language processing (NLP)~\cite{brown2020language,devlin2018bert,liu2023pre,li2021prefix}, numerous studies have been proposed to explore its potential in other domains, such as vision-related and multi-modal scenarios~\cite{chen2022adaptformer,zhou2022conditional,zhou2022learning}. 
VPT~\cite{jia2022visual} takes the first step to visual prompting by adapting vision transformers to downstream tasks with a set of learnable tokens at the model input. Concurrently, VP~\cite{bahng2022exploring} follows a pixel-level perspective to optimize task-specific patches that are incorporated with input images. Although not outperforming full fine-tuning, VP yields an advantage of parameter-efficiency, 
necessitating significantly fewer parameters and a smaller dataset to converge.

Subsequent works explore the properties of VP from different angles. \cite{chen2023understanding} proposed to use different label mapping methods to further tap the potential of VP. \cite{oh2023blackvip} proposes to restrict access to the structure and parameters of the pre-trained model, and puts forward an effective scheme for learning VP under a more realistic setting. In addition,~\cite{chen2023visual} explores the use of VP as a means of adversarial training to improve the robustness of the model, however, their method is limited to the in-domain setting, which is contrary to the original cross-domain transfer intention of VP. It is worth noting that current works on VP are all focused on scenarios where the pre-trained source model is a standard model, and no work has yet investigated the characteristics of VP when originating from a robust source model.

\noindent{\bf Robust Model and Adversarial Training.}
\cite{goodfellow2014explaining} firstly proposes the concept of adversarial examples, in which they add imperceptible perturbations to original samples, fooling the most advanced Deep Neural Networks (DNNs) of that time. Since then, an arms race of attack and defense has begun~\cite{chakraborty2018adversarial,ilyas2018black}. Among the array of defense techniques, adversarial training stands out as the quintessential heuristic method
and has spawned a range of variant techniques~\cite{tramer2017ensemble,tramer2019adversarial}. 
It is a consensus that adversarially trained models possess more complex decision boundaries~\cite{madry2018towards,croce2020minimally}. This complexity arises from adversarial training compelling the classifier to expand the representation of a single class to encompass both clean samples and their adversarially perturbed counterparts~\cite{madry2018towards}. 
Meanwhile, it is broadly recognized that although robust models may exhibit adversarial robustness, this typically comes at the expense of reduced standard accuracy~\cite{chan2019jacobian,gowal2020uncovering}. 

Numerous studies have delved into the above trade-off phenomenon. \cite{tsipras2018robustness} proposes that there may exist an inherent tension between the goal of adversarial robustness and that of standard generalization, discovering that this phenomenon is a consequence of robust classifiers learning fundamentally different feature representations than standard classifiers. \cite{allen2022feature} 
points out that adversarial training could guide models to remove mixed features, leading to purified features (Feature Purification), thus visually conforming more to human perception. 
Moreover, some works believe that the trade-off can be avoided \cite{pang2022robustness} and provide experimental or theoretical proofs. 
There is yet a perfect explanation for this phenomenon. 

Current research indicates that VP is effective in learning and transferring knowledge from standard source models.
However, the inheritance of the unique properties of robust source models by VP remains an area that has yet to be explored.
In this paper, we explore this hitherto unexplored territory for the first time and present the first solution to the negative effects observed in this scenario.
\section{Preliminaries}
{\bf Standard and Adversarial Training.}
In standard classification tasks, the main goal is to enhance standard accuracy, focusing on a model’s ability to generalize to new data that come from the same underlying distribution. The aim here can be defined as achieving the lowest possible expected loss:

\begin{equation}\label{eq:1}
\min_{\theta} \mathbb{E}_{(x,y) \sim D} [\mathcal{L}(x,\theta,y)]
\end{equation}
where $(x,y) \sim D$ represents the training data $x$ and its label $y$ sampled from a particular underlying distribution $D$, 
and $\mathcal{L}$ represents the training loss, i.e., the cross-entropy loss.

After \cite{goodfellow2014explaining} firstly introduce the concept of adversarial training, some subsequent works further refine this notion by formulating a min-max problem, where the goal is to minimize classification errors against an adversary that add perturbations to the input to maximize these errors:

\begin{equation}\label{eq:2}
\min_{\theta} \mathbb{E}_{(x,y) \sim D} [\max_{\delta \in \Delta }\mathcal{L}(x+\delta,\theta,y)]
\end{equation}
where $\Delta$ refers to the set representing the perturbations allowed to be added to the training data $x$ within the maximum perturbation range $\epsilon$, we can define it as a set of $l_{p}$-bounded perturbation, i.e. $\Delta = \{ \delta \in R^{d} \mid \lVert \delta \rVert_{p} \leq \epsilon
  \}$. 

\noindent{\bf Visual Prompt Learning under Robust Models.} 
For a specific downstream dataset, the goal of visual prompting is to learn a prompt that can be added to the data, thus allowing the knowledge of a pre-trained model to be transferred to it. The objective can be formally expressed as follows: 

\begin{align}\label{eq:3}
\min_{\varphi} \, \mathbb{E}_{(x_{t},y_{t}) \sim D_{t}} [\mathcal{L}(\mathcal{M}(f_{\theta^{*}}(\gamma_{\varphi}(x_{t})), y_{t}))] & \\
\text{s.t.} \quad \theta^{*} = \min_{\theta} \, \mathbb{E}_{(x_{s},y_{s}) \sim D_{s}} & [\mathcal{L}(x_{s},\theta,y_{s})]
\end{align}
when the pre-trained model is a robust model, the conditional term in Eq.\ref{eq:3} is changed to:
\begin{equation}\label{eq:4}
\quad \theta^{*} = \min_{\theta} \, \mathbb{E}_{(x_{s},y_{s}) \sim D_{s}} [\max_{\delta \in \Delta }\mathcal{L}(x_{s}+\delta,\theta,y_{s})]
\end{equation}
where $D_{t}$ and $D_{s}$ represent the distribution of the downstream dataset and the source dataset, respectively; $f_{\theta^{*}}(\cdot)$ represents the frozen pre-trained model parameterized by $\theta^{*}$; $\gamma_{\varphi}({\cdot})$, parameterized by $\varphi$, represents the visual prompt that needs to be learned; \(\mathcal{M}(\cdot)\) represents the pre-defined label mapping strategy. It assumes that the dataset used to train the source model typically includes a larger number of classes. Consequently, a subset of dimensions from the source model's final linear layer is selected using a specific strategy, and this subset is employed to create a one-to-one mapping to the classes in the downstream dataset.

\section{Observations Under RSVP}
As mentioned earlier, 
existing works primarily focus on understanding VP in the context of standard models, the unique inheritance characteristics of VP under RSVP, as well as solutions for its specific disadvantages, remain to be explored. In this section, we explore these questions and present our findings. Our attempts to address its specific disadvantages will be discussed in the next section.

\begin{figure}[t]
    \centering
    \begin{subfigure}[b]{\columnwidth}  
        \includegraphics[width=\textwidth]{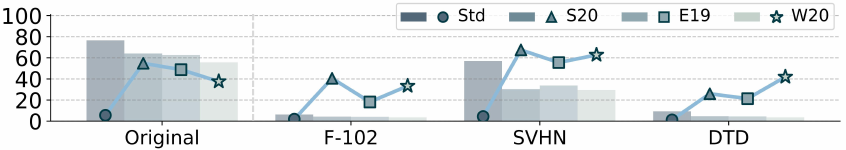}
        \caption{Random Label Mapping}
        \label{fig:sub1}
    \end{subfigure}

    \begin{subfigure}[b]{\columnwidth}  
        \includegraphics[width=\textwidth]{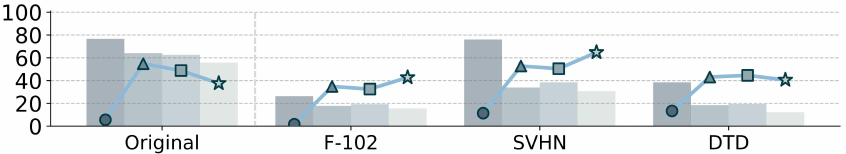}
        \caption{Iterative Label Mapping}
        \label{fig:sub2}
    \end{subfigure}

    \caption{The performance of VP on standard accuracy (histogram) and adversarial accuracy (line chart) when using a standard model or different robust models as the source model. `Original' represents the result on the source dataset without VP.}
    \label{fig:contrast}
\end{figure}

\noindent{\bf Robustness Inheritance of Visual Prompt.} Initially, we investigate the extent to which a source model's robustness transfers to visual prompts. Intuitively, since the source dataset and downstream datasets belong to different domains, and adversarial training is specifically tailored to the source dataset, inheriting robustness for visual prompts appears neither straightforward nor effortless.
We use models from RobustBench \cite{croce2021robustbench}, an open-source benchmark widely used in trustworthy machine learning. Specifically, we select one standard model (referred to as Std) and three robust models trained with ImageNet \cite{deng2009imagenet} under the $l_{\infty}$-norm. The three robust models are referred to as S20 \cite{salman2020adversarially}, E19 \cite{robustness} and W20 \cite{wong2020fast}, respectively. Without loss of generality, we used FGSM (Fast Gradient Sign Method) attack \cite{goodfellow2014explaining} to assess the robustness of each model. For datasets, we use flowers102 (F-102) \cite{nilsback2008automated}, SVHN \cite{netzer2011reading} and DTD \cite{cimpoi2014describing} for this experiment. The results are shown in Figure \ref{fig:contrast}, among which Figure \ref{fig:contrast} (a) and Figure \ref{fig:contrast} (b) represent the results under different label mapping methods, respectively. The bar chart represents the results of standard accuracy, while the line chart represents the results of adversarial accuracy. 'Original' denotes the performance of the source model on its original source dataset without utilizing VP for knowledge transfer.

The bar charts in Figure \ref{fig:contrast} illustrate that visual prompts derived from a standard source model do not exhibit robustness. In contrast, visual prompts trained with robust source models demonstrate markedly improved robustness compared to their standard-trained counterparts. 
Moreover, we observe that a given source model yields varying outcomes across different downstream datasets. Similarly, for a specific downstream dataset, the results differ when using various source models. 

\noindent{\bf Generalization Ability Encountered Degradation.} 
We further explore the disparities in standard accuracy between SSVP and RSVP in various downstream datasets. The line charts in Figure \ref{fig:contrast} show a decrease in the generalization performance of RSVP compared to SSVP, reflecting the performance trend (i.e., the generalization-robustness trade-off) observed in the source model itself.

Additionally, we observe no clear relationship between the performance gaps of various robust source models and the RSVP performance disparities derived from them. This indicates that improving the robustness or generalization ability of the source model does not necessarily lead to corresponding enhancements in RSVP performance. In fact, such attempts may be ineffective or even detrimental. Thus, a customized strategy is essential for RSVP to increase its generalization ability while maintaining or potentially increasing its robustness. Our proposed PBL represents an initial foray into addressing this challenge.

\begin{figure}[t]
  \centering
   \includegraphics[width=\columnwidth]{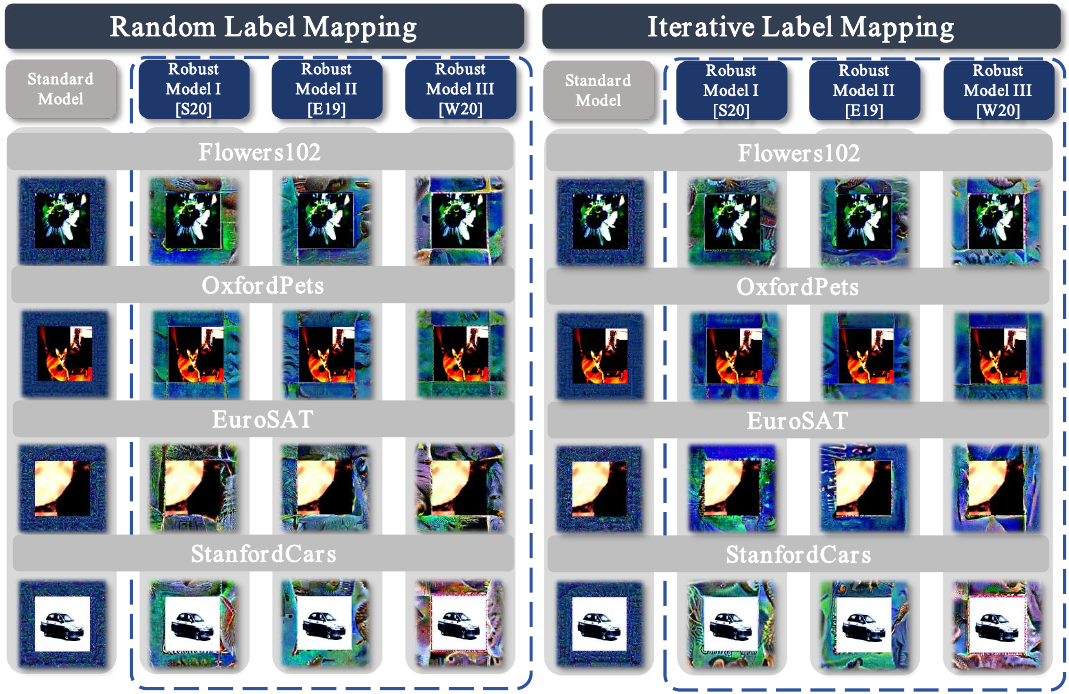}

   \caption{Visual representation of SSVP (columns 1 \& 5) and RSVP (columns 2-4 \& 6-8) obtained during a certain training period. For SSVP, only meaningless noise can be observed, while for RSVP, we get a representation consistent with human perception.}
   \label{fig:rbt_vp}
\end{figure}
\noindent{\bf Visual Representation of Visual Prompt under Robust Models.}
All current VP-related works focus on the case of SSVP. Under this setting, as shown in columns 1 and 5 of Figure \ref{fig:rbt_vp}, the learned prompt appears to be random noise without any meaningful visual representations. 
In this work, we visualize RSVP and find, surprisingly, that RSVP (as shown in columns 2-4 and 5-8 of Figure \ref{fig:rbt_vp}) exhibits visual representations that align well with human perception—possessing distinct shapes, textures, or recognizable objects (additional examples are provided in the supplemental materials). This phenomenon consistently occurs across various robust models, label mapping methods, and datasets.

The above phenomenon provides potential insights into how RSVP inherits robustness from source models. Referring to Eq.\ref{eq:3} and Eq.\ref{eq:4}, a VP with learnable parameters takes an original image as input and generates an image-like output (hereafter referred to as a trainable image). This trainable image is then fed into the pre-trained source model for prediction. If VP and the original image are considered as a unified entity, the process of training VP can essentially be interpreted as calculating and back-propagating the loss gradient with respect to a subset of the input image pixels.
Previous works \cite{tsipras2018robustness,allen2022feature} suggest that adversarial robustness and standard generalization performance might be at odds with each other, which is attributed to the fact that the feature representations of standard and robust models are fundamentally different. To illustrate, in the absence of a VP, when one calculates the loss gradient with respect to the input image pixels (this operation can highlight the input features that significantly influence the loss and hence the model's prediction), it becomes evident upon visualization that robust models develop representations that are more aligned with prominent data features and human perception, which is consistent with the traits exhibited by RSVP.


\section{Attempts to Mitigate the Trade-Off}
The above findings indicate that while RSVP inherits the robustness of the source model, it also experiences a comparable decline in standard accuracy, much like the source model. This limitation significantly constrains its practical applicability.
In this section, 
we introduce the Prompt Boundary Loosening (PBL) as a solution to address the shortcomings. In short, our objectives can be summarized into two main aspects.
\textbf{Obj-1}: Achieving lightweight yet effective robust transfer that balances robustness and generalization, while avoiding the extensive time and computational demands typical of adversarial training;
and \textbf{Obj-2}: Naturally adapting to the inherent settings of VPs, where the source model remains frozen and label mapping is used for adaptation—meaning we aim for a solution that is independent of both the source model and the label mapping strategy.


Referring to Eq.\ref{eq:3} and Eq.\ref{eq:4}, each input image from the target downstream dataset is first processed by RSVP and then passed through the source model, resulting in a predicted probability \( f_{\theta^{*}}(\gamma_{\varphi}(x_{t})) \), which matches the dimensionality of the source dataset. Subsequently, the predefined label mapping method \( \mathcal{M}(\cdot) \) is applied to derive the final predicted probability for the target dataset.

In the RSVP scenario, the source model is an adversarially trained robust model with a more complex decision boundary compared to a standard-trained model (see Section \ref{sec:RelatedWork}). However, within the VP learning pipeline, the decision boundary of the frozen source model remains fixed, which significantly increases the learning difficulty of RSVP.
It might be assumed that enhancing RSVP's ability to learn from a complex decision boundary could be achieved by scaling up the prompt to introduce more learnable parameters. However, existing research \cite{bahng2022exploring} indicates that such scaling provides only marginal improvements to the performance, and beyond a certain point, it may even negatively impact the effectiveness of the prompt.
Motivated by the aforementioned insights and observations, we introduce PBL as an initial step towards advancing the functionality of RSVP. 

Specifically, PBL can be defined as a function $\mathcal{Q}(\cdot)$, which receives the output of the source model $f_{\theta^{*}}(\gamma_{\varphi}(x_{t}))$ and a loosening factor $\mathcal{T}$ as inputs, then randomly combines the elements of $f_{\theta^{*}}(\gamma_{\varphi}(x_{t}))$ according to $\mathcal{T}$ to output an intermediate vector with a smaller dimension than the original output, then do the label mapping step $\mathcal{M}(\cdot)$ on this vector to get the final prediction for the target downstream dataset. By formalizing the objective function with PBL, we get:

\begin{equation}\label{eq:5}
\begin{aligned}
& \min_{\varphi} \, \mathbb{E}_{(x_{t},y_{t}) \sim D_{t}} [\mathcal{L_{PBL}}(\mathcal{M}(\mathcal{Q}(f_{\theta^{*}}(\gamma_{\varphi}(x_{t})), \mathcal{T}), y_{t}))] \\
& \text{s.t.} \quad \theta^{*} = \min_{\theta} \, \mathbb{E}_{(x_{s},y_{s}) \sim D_{s}}  [\max_{\delta \in \Delta }\mathcal{L}(x_{s}+\delta,\theta,y_{s})]
\end{aligned}
\end{equation}

We assume that the dimension of the output of the source model is $n$, and record the original output $f_{\theta^{*}}(\gamma_{\varphi}(x_{t}))$ as a vector $V = (v_1, v_2, ..., v_n)$. We deal with $n/\mathcal{T}$ elements at once and divide $V$ into $\mathcal{T}$ parts, each of which is marked as: 

\begin{equation}\label{eq:6}
\begin{aligned}
V_i = (v_{(i-1)n/\mathcal{T} + 1}, v_{(i-1)n/\mathcal{T} + 2}, ..., v_{in/\mathcal{T}}), & \\
i = 1, 2, ..., \mathcal{T}
\end{aligned}
\end{equation}

Suppose the intermediate vector is called $\mathcal{I}$, its $i^{th}$ element is the maximum value in the $i^{th}$ partition of $V$, i.e., $I_i = \max(V_i)$, which means taking the maximum confidence score in the current merged block as a representative value.
$\mathcal{I}$ can be expressed as: 

\begin{equation}\label{eq:7}
\mathcal{I} = (\max(V_1), \max(V_2), ..., \max(V_T))
\end{equation}

The core intuition behind the intermediate vector \(\mathcal{I}\) is to fully leverage the knowledge the source model has acquired from the source dataset during the initial stage of knowledge transfer (see Section \ref{exp}).
In addition, the looser decision area increases the quality of label mapping, thereby reducing the prediction difficulty for the downstream dataset (see Section \ref{exp}). Finally, \(\mathcal{I}\) can be used to map the downstream dataset and generate the final predictions:

\begin{equation}\label{eq:8}
\begin{aligned}
&\mathcal{L_{PBL}}(\mathcal{M}(\mathcal{Q}(f_{\theta^{*}}(\gamma_{\varphi}(x_{t})), \mathcal{T}), y_{t})) \\
& = \mathcal{L_{PBL}}(\mathcal{M}(\mathcal{Q}(V, \mathcal{T}), y_{t})) \\
& = \mathcal{L_{PBL}}(\mathcal{M}(\mathcal{I}, y_{t}))
\end{aligned}
\end{equation}


Note that when applying VP to data from the same class in the downstream dataset, the source model may produce varying predictions, with the highest prediction probability corresponding to different classes. Additionally, some individual data points may display multiple high-confidence scores.
The loosening factor \(\mathcal{T}\) in PBL formally relaxes the decision boundary of \(f_{\theta^{*}}(\cdot)\), thereby reducing prediction difficulty and mitigating the low accuracy caused by the aforementioned phenomenon. At the same time, it preserves and utilizes the intricate decision boundary of the source model, ensuring that the robustness transferred from the source model is effectively retained.
We find that PBL is highly compatible with existing label mapping methods and can serve as a seamless, plug-and-play enhancement to enable the training of more effective VPs.

\begin{table*}[ht]
\centering
\small
\setlength\tabcolsep{3pt} 
\resizebox{\textwidth}{!}{ 
\begin{tabular}{c|c|cc|cc|cc|cc|cc|cc|cc|cc}
\toprule
\multirow{2}{*}{LM} & \multirow{2}{*}{Dataset} & \multicolumn{4}{c|}{ResNet18} & \multicolumn{4}{c|}{ResNet50} & \multicolumn{4}{c|}{Wide-ResNet50-2} & \multicolumn{4}{c}{ViT-S} \\
\cmidrule(lr){3-6} \cmidrule(lr){7-10} \cmidrule(lr){11-14} \cmidrule(lr){15-18} 
& & Adv. \footnotesize(w/o) & Adv. \footnotesize(w) & Std. \footnotesize(w/o) & Std. \footnotesize(w) & Adv. \footnotesize(w/o) & Adv. \footnotesize(w) & Std. \footnotesize(w/o) & Std. \footnotesize(w) & Adv. \footnotesize(w/o) & Adv. \footnotesize(w) & Std. \footnotesize(w/o) & Std. \footnotesize(w) & Adv. \footnotesize(w/o) & Adv. \footnotesize(w) & Std. \footnotesize(w/o) & Std. \footnotesize(w) \\
\midrule
\multirow{8}{*}{\rotatebox[origin=c]{90}{Random-LM} } 
& F-102 & \textbf{33.33\%} & 32.79\% & 5.24\% & \textbf{7.43\%} & 40.57\% & \textbf{46.02\%} & 4.30\% & \textbf{4.63\%} & 19.33\% & \textbf{45.92\%} & 4.83\% & \textbf{5.24\%} & \textbf{39.17\%} & 31.55\% & 7.88\% & \textbf{8.36\%}  \\
& DTD & 26.47\% & \textbf{39.36\%} & 4.02\% & \textbf{5.56\%} & 25.97\% & \textbf{37.17\%} & 4.55\% & \textbf{6.68\%} & \textbf{45.16\%} & 42.14\% & 5.50\% & \textbf{8.33\%} & 37.93\% & \textbf{39.88\%} & 8.22\% & \textbf{9.93\%} \\
& SVHN & 71.67\% & \textbf{74.21\%} & 32.23\% & \textbf{34.28\%} & \textbf{67.50\%} & 59.08\% & 30.18\% & \textbf{34.70\%} & 52.62\% & \textbf{58.16\%} & 35.50\% & \textbf{38.75\%} & \textbf{44.04\%} & 41.91\% & 44.43\% & \textbf{45.23\%} \\
& G-RB & 53.03\% & \textbf{78.71\%} & 12.42\% & \textbf{13.84\%} & 74.35\% & \textbf{77.16\%} & 11.95\% & \textbf{14.11\%} & 75.38\% & \textbf{80.99\%} & 15.08\% & \textbf{17.87\%} & 58.21\% & \textbf{61.66\%} & 19.38\% & \textbf{22.41\%} \\
& E-Sat & 46.45\% & \textbf{47.70\%} & 50.72\% & \textbf{53.46\%} & 43.59\% & \textbf{50.91\%} & 53.05\% & \textbf{57.78\%} & 42.46\% & \textbf{52.73\%} & 54.23\% & \textbf{56.31\%} & 28.94\% & \textbf{30.79\%} & \textbf{62.89\%} & 62.44\% \\
& O-Pets & 5.83\% & \textbf{14.75\%} & 3.27\% & \textbf{4.99\%} & 14.39\% & \textbf{16.57\%} & 3.60\% & \textbf{4.93\%} & 18.85\% & \textbf{24.60\%} & 3.33\% & \textbf{6.76\%} & \textbf{14.48\%} & 14.04\% & 7.90\% & \textbf{8.42\%} \\
& CI-100 & 75.07\% & \textbf{77.13\%} & 3.53\% & \textbf{4.95\%} & 66.26\% & \textbf{72.20\%} & 4.97\% & \textbf{5.16\%} & \textbf{77.77\%} & 74.06\% & 3.79\% & \textbf{5.61\%} & 53.63\% & \textbf{57.79\%} & 5.78\% & \textbf{5.97\%} \\
& S-Cars & 13.04\% & \textbf{13.11\%} & 0.57\% & \textbf{0.76\%} & 6.66\% & \textbf{20.37\%} & 0.56\% & \textbf{0.67\%} & 28.81\% & \textbf{32.69\%} & 0.73\% & \textbf{0.83\%} & \textbf{24.53\%} & 11.11\% & 0.66\% & \textbf{0.78\%} \\
\midrule
\multirow{8}{*}{\rotatebox[origin=c]{90}{Iterative-LM} } 
& F-102 & 40.65\% & \textbf{44.66\%} & 18.88\% & \textbf{22.82\%} & 34.36\% & \textbf{34.86\%} & 17.70\% & \textbf{22.45\%} & 34.38\% & \textbf{37.25\%} & 19.53\% & \textbf{20.71\%} & 21.59\% & \textbf{26.54\%} & 14.29\% & \textbf{17.74\%} \\
& DTD & 41.11\% & \textbf{44.03\%} & 15.96\% & \textbf{18.79\%} & 43.09\% & \textbf{50.87\%} & 18.38\% & \textbf{20.45\%} & \textbf{54.28\%} & 50.14\% & 20.04\% & \textbf{21.45\%} & 23.78\% & \textbf{25.81\%} & 19.98\% & \textbf{22.28\%} \\
& SVHN & 61.67\% & \textbf{65.85\%} & 34.47\% & \textbf{35.44\%} & 52.76\% & \textbf{57.77\%} & 33.85\% & \textbf{34.96\%} & 52.85\% & \textbf{54.32\%} & 36.67\% & \textbf{37.60\%} & 40.28\% & \textbf{48.81\%} & 44.38\% & \textbf{45.86\%} \\
& G-RB & \textbf{68.96\%} & 67.92\% & 17.47\% & \textbf{20.24\%} & 74.42\% & \textbf{75.15\%} & 17.64\% & \textbf{19.46\%} & 62.23\% & \textbf{64.82\%} & 18.50\% & \textbf{19.26\%} & 54.28\% & \textbf{60.62\%} & 21.13\% & \textbf{23.12\%} \\
& E-Sat & 41.21\% & \textbf{42.13\%} & 59.20\% & \textbf{61.83\%} & 47.32\% & \textbf{47.36\%} & 58.12\% & \textbf{63.10\%} & \textbf{53.87\%} & 53.68\% & 55.59\% & \textbf{60.72\%} & \textbf{30.02\%} & 29.26\% & 62.17\% & \textbf{64.26\%} \\
& O-Pets & 32.84\% & \textbf{35.55\%} & 16.60\% & \textbf{23.00\%} & \textbf{38.53\%} & 38.15\% & 27.15\% & \textbf{33.74\%} & \textbf{38.25\%} & 37.18\% & 34.21\% & \textbf{36.17\%} & 22.92\% & \textbf{23.84\%} & \textbf{42.44\%} & 41.31\% \\
& CI-100 & 65.34\% & \textbf{68.99\%} & 11.60\% & \textbf{12.81\%} & \textbf{60.80\%} & 59.19\% & 11.51\% & \textbf{12.70\%} & 64.69\% & \textbf{64.71\%} & 10.97\% & \textbf{12.28\%} & \textbf{50.22\%} & 47.97\% & 11.25\% & \textbf{13.05\%} \\
& S-Cars & 20.26\% & \textbf{25.00\%} & 1.90\% & \textbf{2.29\%} & 24.44\% & \textbf{27.22\%} & 1.68\% & \textbf{2.10\%} & \textbf{33.57\%} & 33.14\% & 1.78\% & \textbf{2.23\%} & 14.89\% & \textbf{15.10\%} & 1.75\% & \textbf{1.80\%} \\
\bottomrule
\end{tabular}}
\caption{Performance of our proposed Prompt Boundary Loosening (PBL) under RSVP setting over eight downstream datasets and four pre-trained robust source models (ResNet-18, ResNet-50, Wide-ResNet50-2 and ViT-S trained on ImageNet). Adv. (w/o) and Std. (w/o) means Adversarial Accuracy and Standard Accuracy without using PBL, while Adv. (w) and Std. (w) means Adversarial Accuracy and Standard Accuracy when using PBL. The better outcomes are marked in bold.}
\label{tab:3}
\end{table*}
\section{Experiments}
\label{exp}
In this section, we empirically demonstrate the effectiveness of PBL in the inheritance of both robustness and standard accuracy under RSVP. Additionally, we explore the characteristics of PBL from multiple perspectives and provide valuable insights into its inheritance mechanisms.


\subsection{Experimental Settings}
\newcommand{\mysubsection}[1]{\noindent $\bullet$ \textbf{#1}}
\mysubsection{Models and Datasets.} We use two types of source model: Standard Source Model and Robust Source Model, both of which include four types of model pre-trained on ImageNet-1K. For Standard Source Model, we use the pre-trained models from \textit{torch} and \textit{timm} \cite{rw2019timm}, while for Robust Source Model, we use the pre-trained models from \textit{RobustBench} \cite{croce2021robustbench} same as in Figure \ref{fig:contrast}. All the models we use are pre-trained on ImageNet.
We consider 8 downstream datasets: Flowers102 (F-102) \cite{nilsback2008automated}, DTD \cite{cimpoi2014describing}, GTSRB (G-RB) \cite{stallkamp2011german}, SVHN \cite{netzer2011reading}, EuroSAT (E-Sat) \cite{helber2019eurosat}, OxfordPets (O-Pets) \cite{parkhi2012cats}, StanfordCars (S-Cars) \cite{krause20133d} and CIFAR100 (CI-100) \cite{krizhevsky2009learning}.

\mysubsection{Evaluations and Baselines.}
Without lose of generality, we consider two widely used label mapping strategies \cite{chen2023understanding}: Random Label Mapping (RLM) and Iterative Label Mapping (ILM). RLM refers to randomly matching the labels of the source dataset to those of the target dataset before training, while ILM refers to re-matching the labels of the source dataset to those of the target dataset according to the model prediction after each iteration, so as to make full use of the training dynamics of VP. For each LM-Dataset-Model combination, we explore the standard accuracy (Std. Acc) as well as the adversarial accuracy (Adv. Acc) with or without PBL. FGSM~\cite{goodfellow2014explaining} is used as the attack method. Note that adversarial attacks are only performed on data that the model initially classifies correctly, with the goal of causing the model’s correct prediction to become incorrect (i.e., $\text{Std. Acc} = \frac{\# \text{Ori. Correct Samples}}{\# \text{All Samples}};
\text{Adv. Acc} = \frac{\# \text{Adv. Correct Samples}}{\# \text{Ori. Correct Samples}}
$).


In our experiments, we will show the effectiveness of PBL under different source models and datasets. Also, we will explore the characteristics of PBL from multiple perspectives. Furthermore, we will investigate the impact of additional adversarial training under RSVP, analyzing the results in terms of standard and adversarial accuracy, time usage, and computational resource consumption.

\subsection{PBL brings benefits to RSVP}
Table \ref{tab:3} shows the main results under the RSVP scenario. We consider the combinations of 8 different datasets, 4 different source model architectures and 2 different LM methods. The first two columns for each model architecture shows the capability of PBL in inheriting robustness, while the latter two columns show its effectiveness in improving the generalization performance. 

\begin{figure}[t]
    \centering
    \begin{subfigure}[b]{0.75\columnwidth}
        \centering
        \includegraphics[width=\columnwidth]{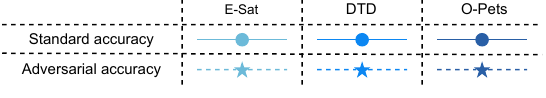}
        \label{fig:first-image}
    \end{subfigure}
    \begin{subfigure}[b]{0.4\columnwidth}
        \centering
        \includegraphics[width=\columnwidth]{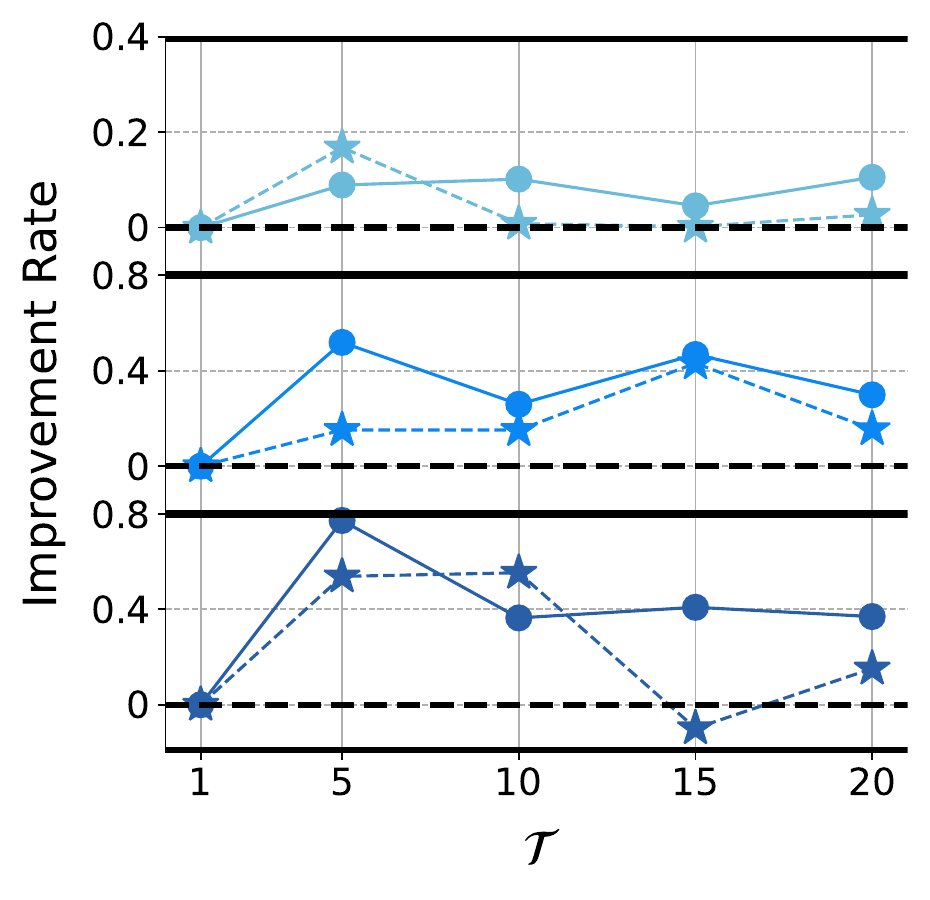}
        \caption{RLM}
        \label{fig:random-mapping}
    \end{subfigure}
    \begin{subfigure}[b]{0.4\columnwidth}
        \centering
        \includegraphics[width=.98\columnwidth]{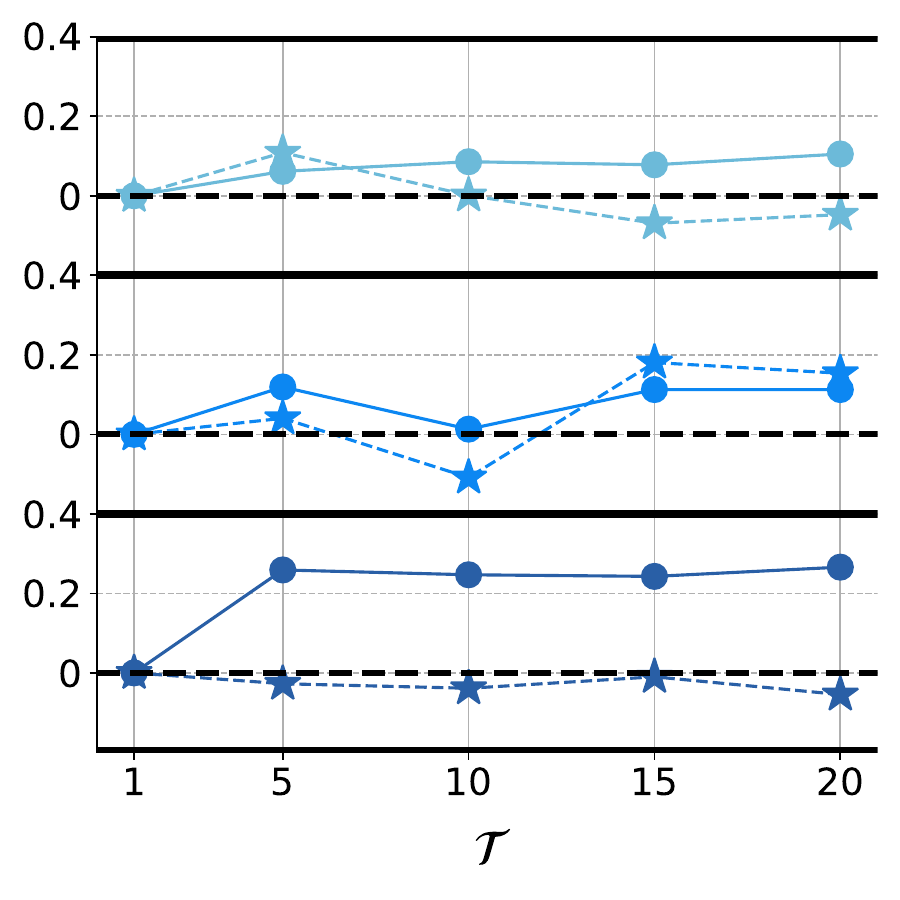}
        \caption{ILM}
        \label{fig:iterative-mapping}
    \end{subfigure}
    \caption{The performance improvement of PBL in EuroSAT, DTD and OxfordPets at different loosening factors $\mathcal{T}$, the standard accuracy is represented by solid lines and circles, while the adversarial accuracy is represented by dotted lines and asterisks.}
    \label{fig:diff-T}
\end{figure}

The results consistently show that VP achieves improved generalization across all downstream datasets. Additionally, the robustness of the source model is effectively inherited and, in some instances, even substantially enhanced.
Specifically, with ResNet50 as the source model, Std. Acc of E-Sat is improved by 4.73\% under RLM and 4.98\% under ILM. As for robustness, 
for instance, when the source model and LM methods are ResNet18 and RLM, the Adv. Acc of DTD increases by 12.89\% and the Adv. Acc of OxfordPets increases by 8.92\%. 
Moreover, our findings indicate that \textit{superior label mapping methods (e.g., ILM over RLM) can enhance standard accuracy but do not guarantee that VP can better inherit the robustness of the source model.}
For instance, with ResNet18 as source model, when not utilizing PBL, robustness of E-Sat drops from 46.45\% under RLM to 41.21\% under ILM—a reduction of 5.24\%. Similarly, robustness of CI-100 decreases from 75.07\% with RLM to 65.34\% with ILM.
In most cases, PBL generally enables VP to better inherit robustness of the source model, regardless of the label mapping method applied. Therefore, it can be regarded as a plug-and-play component that perfectly aligns with the characteristics of the visual prompting process.
\begin{figure}
    \centering
    \begin{subfigure}[b]{\columnwidth}    
        \includegraphics[width=\textwidth]{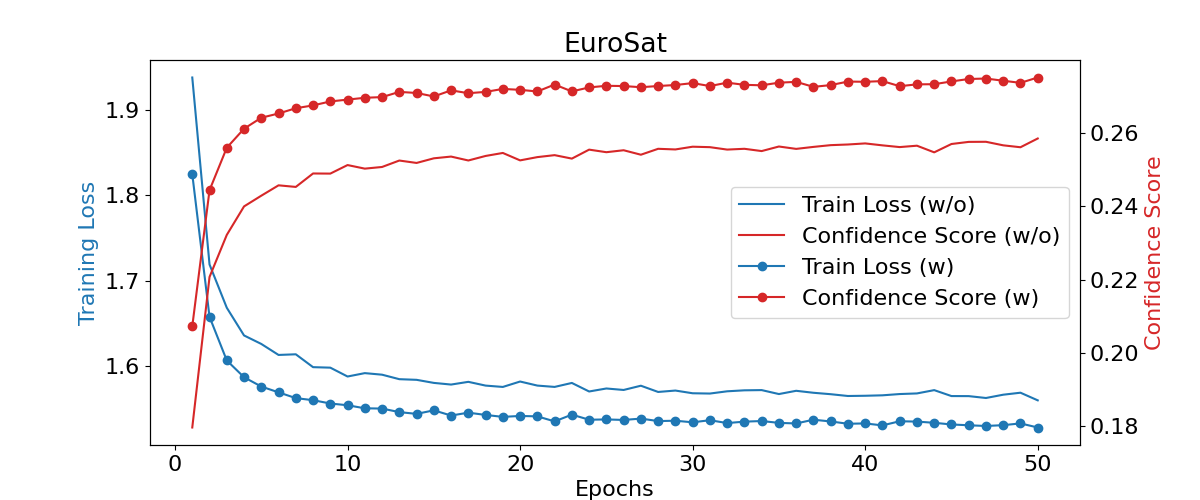}
        \caption{Training Dynamics of EuroSat}
        \label{fig:eurosat}
    \end{subfigure}
    \hfill
    \begin{subfigure}[b]{\columnwidth}
        \includegraphics[width=\textwidth]{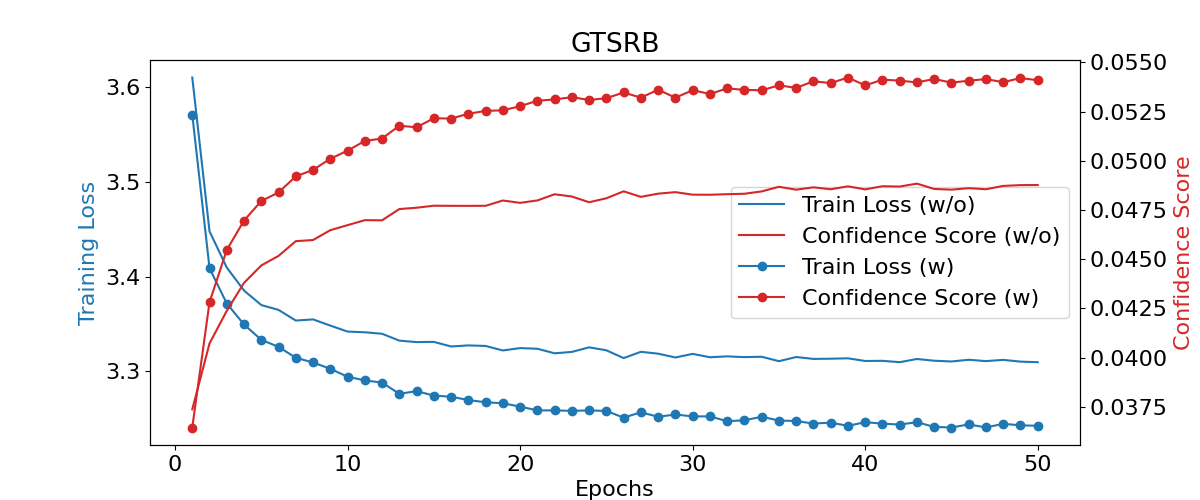}
        \caption{Training Dynamics of GTSRB}
        \label{fig:gtsrb}
    \end{subfigure}
    \caption{The training dynamics for the EuroSat and GTSRB datasets during the first 50 epochs utilizing RLM. PBL proves beneficial in the early stage of training.}
    \label{fig:dynamics}
\end{figure}

\begin{table}[h]
  \centering
  \small
  \resizebox{\linewidth}{!}{
  \begin{tabular}{c|c|c|c|c|c}
      \toprule
      Dataset    & Perf. & w/o. PBL & w/o. PBL+AT & w. PBL & w. PBL+AT \\
      \midrule
      \multirow{2}{*}{F-102} & Std. & 17.70\% & 16.16\% & 22.45\% & 19.20\% \\
      & Adv. & 34.36\% & 53.27\% & 34.86\% & 52.43\% \\
      \midrule
      \multirow{2}{*}{DTD} & Std. & 18.38\% & 17.61\% & 20.45\% & 19.27\% \\
      & Adv. & 43.09\% & 51.68\% & 50.87\% & 51.23\% \\
      \midrule
      \multirow{2}{*}{O-Pets} & Std. & 27.15\% & 24.83\% & 33.74\% & 31.53\% \\
      & Adv. & 38.53\% & 37.10\% & 38.15\% & 38.14\% \\
      \bottomrule
  \end{tabular}}
  \caption{Result of using four different strategy combinations in different datasets. AT can improve robustness in some cases, however, sometimes it can not bring considerable gain but will consume more resources. In contrast, PBL can improve standard accuracy while maintaining robustness regardless of whether AT is utilized or not.}
  \label{tav:adv_train}
\end{table}

\begin{figure}[h]
  \centering
  \includegraphics[width=\linewidth]{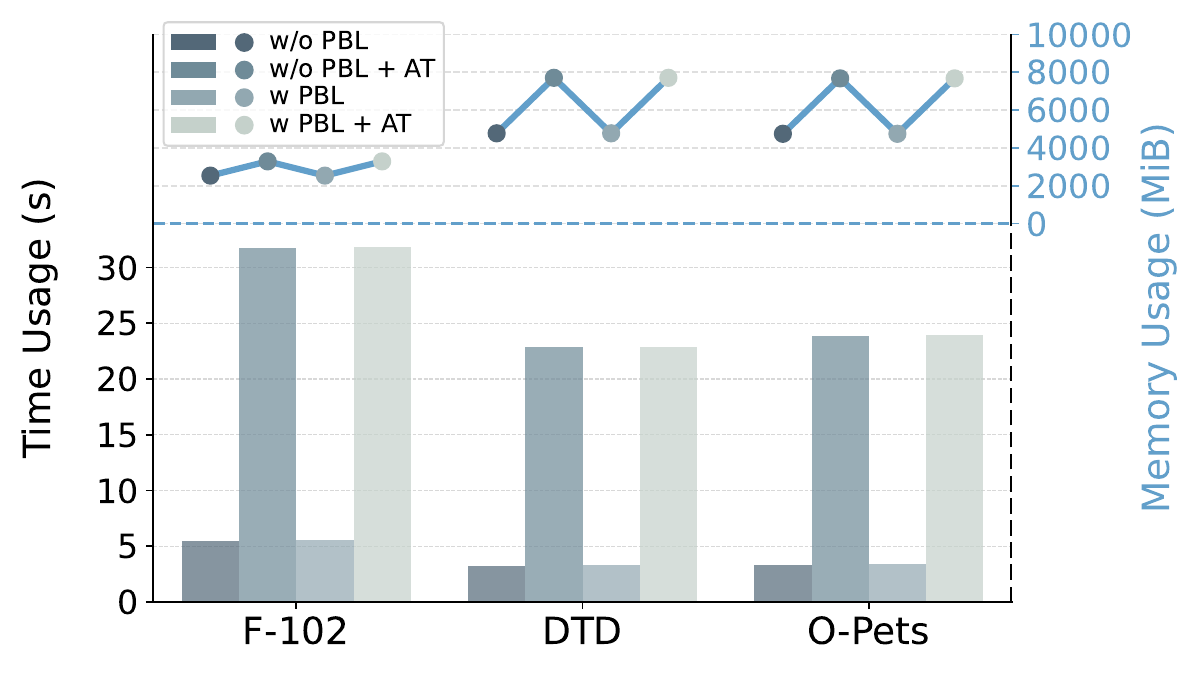}
  \caption{Time usage and resource consumption under different combinations of PBL and AT. The bar chart represents time usage while the line chart represents the computing resource consumption. Results are mean values per epoch.}
  \label{fig:adv_train}
  \vspace{-5mm}
\end{figure}

As mentioned before, the computation of adversarial accuracy (Adv. Acc) presupposes the model's correct initial classification of a sample—we only attempt an attack on samples that the model has accurately identified pre-attack.
Hence, \textit{due to the generalization performance enhancement brought by applying PBL, employing PBL typically results in a larger set of samples subject to attack. 
Therefore, when using PBL, it becomes more challenging to preserve or enhance the Adv. Acc of RSVP.} Considering this, the simultaneous improvement in generalization and robustness brought by PBL to RSVP becomes even more significant.


\subsection{Understanding of PBL} 

\mysubsection{General advantages at different loosening factor $\mathcal{T}$.} 
Without lose of generality, we set $\mathcal{T}$ to five values between 1 and 20 on EuroSAT, DTD and OxfordPets with ResNet50 as the source model. As shown in Figure \ref{fig:diff-T}, the value of $\mathcal{T}$ = 1 in the x-axis is set as the zero point to indicate the baseline performance without PBL. Performance at different $\mathcal{T}$s is measured as the improvement rate relative to this baseline.

We can find that regardless of the loosening factor value, PBL consistently yields substantial gains in standard accuracy across the board. Specifically, PBL enhances standard accuracy by approximately 10\% across all $\mathcal{T}$ setups on EuroSAT. With DTD, employing RLM as the label mapping method typically results in a 40\% increase, while OxfordPets sees a peak improvement of around 80\%.
In addition, adversarial accuracy remains stable across different \(\mathcal{T}\) values, with notable improvements at specific points. For instance, using RLM, adversarial accuracy on E-SAT, DTD, and O-Pets increases by up to 20\%, 40\%, and 50\%, respectively, in the most extreme cases.

It is worth noting that different LM methods exhibit a consistent trend in standard accuracy gains across varying $\mathcal{T}$s. 
One possible explanation is that different LM methods may tap into specific phases of the VP training dynamics, including initialization and subsequent updates, to enhance overall performance. Specifically, RLM sets the mapping at beginning and maintains it throughout later iterations, making it dependent solely on the quality of the initialization. ILM continuously revises its mapping sequence post-initialization (which can be seen as a re-initialization), capitalizing on the evolving training dynamics of VP. Meanwhile, PBL helps to pre-define a dynamic initialization for each training iteration from the potential distribution, enhancing the default settings and thereby improving the learning efficiency and efficacy of different LM methods.  

To validate this hypothesis, Figure \ref{fig:dynamics} illustrates the training dynamics for two datasets. From the outset, the less complex decision boundary enables easier transfer of source domain knowledge, resulting in a higher initial average confidence score and lower training loss compared to the non-PBL setup. This advantage is sustained or even amplified during the subsequent training process, highlighting the superior performance facilitated by PBL in the initialization phase.

\begin{table}[h!]
  \centering
  \resizebox{\columnwidth}{!}{
  \begin{tabular}{c|c|cc|cc|cc}
    \toprule
    \multirow{2}{*}{LM} & \multirow{2}{*}{Dataset} & \multicolumn{2}{c}{ResNet18} & \multicolumn{2}{c}{ResNet50} & \multicolumn{2}{c}{ViT-S}  \\
    \cmidrule(lr){3-4} \cmidrule(l){5-6} \cmidrule(l){7-8}
    & & {Std. Acc (w/o)} & {Std. Acc (w)} & {Std. Acc (w/o)} & {Std. Acc (w)} & {Std. Acc (w/o)} & {Std. Acc (w)} \\
    \midrule
    \multirow{3}{*}{\rotatebox[origin=c]{90}{RLM}} & f-102 & 12.02\% & \textbf{13.28\%} & 9.83\% & \textbf{12.06\%} & \textbf{61.39\%} & 60.33\% \\
    & gtsrb & 47.14\% & \textbf{49.05\%} & 45.67\% & \textbf{46.83\%} &58.16\% & \textbf{60.22\%} \\
    & C-100 & 9.95\% & \textbf{11.36\%} & 9.61\% & \textbf{10.82\%} & 29.83\% & \textbf{31.35\%} \\
    \midrule
    \multirow{3}{*}{\rotatebox[origin=c]{90}{ILM}} & f-102 & 29.03\% & \textbf{30.82\%} & 26.23\% & \textbf{26.67\%} & 77.51\% & \textbf{79.66\%} \\
    & gtsrb & 52.86\% & \textbf{54.22\%} & 53.94\% & \textbf{55.61\%} &\textbf{60.96\%} & 60.53\% \\
    & C-100 & 25.08\% & \textbf{27.34\%} & 38.87\% & \textbf{40.50\%} &34.19\% & \textbf{38.45\%} \\
    \bottomrule
  \end{tabular}}
  \caption{Comparison of standard accuracy (Std. Acc.) when using (w) and without using (w/o) PBL under SSVP.}
  \label{tab:ssvp}
\end{table}

\mysubsection{PBL brings benefits to SSVP.}
It would be undesirable to observe an improvement in standard accuracy for RSVP alone if it is not accompanied by similar performance in SSVP, as this would limit the practicality of PBL. To verify the actual impact, we further conduct an experiment to evaluate PBL's performance with SSVP, with the expectation that PBL would not adversely affect generalization performance. As shown in Table \ref{tab:ssvp}, we are pleased to observe that PBL not only significantly improves standard and adversarial accuracy in the RSVP context but also enhances standard accuracy under SSVP—an additional benefit, albeit not the primary objective of PBL. This highlights PBL's versatility as a technique for improving VP performance across various source model types.

\mysubsection{The intolerability of adversarial training for VP.}  
We further investigate the efficacy of additional adversarial training for RSVP. It is worth noting that the standard accuracy for RSVP is already significantly lower than that for SSVP, as a trade-off for robustness. Therefore, applying additional adversarial training to RSVP could further exacerbate the decline in standard accuracy. While this approach may enhance robustness, a model that is robust but lacks generalization ability is meaningless.

In Table \ref{tav:adv_train} and Figure \ref{fig:adv_train}, we assess the impact of PBL and Adversarial Training (AT). In this experiment, for comparative fairness, AT is done on VP while the source model remains frozen. Our analysis encompasses standard and adversarial accuracy, as well as average time usage and computing resource consumption over 200 training epochs, under four distinct combinations of PBL and AT. 
As shown in Table \ref{tav:adv_train}, while adversarial training alone enhances RSVP's robustness (see columns 1 \& 2), it notably compromises standard accuracy. 
Even in some cases, e.g, with DTD and OxfordPets as target datasets, adversarial training not only leads to a reduction in standard accuracy but also offers negligible robustness gains (see columns 2 \& 3), while significantly increasing computational resource consumption ($ \simeq 1.5 \times $) and time usage ($ \simeq 6 \times $), which is intolerable.
In contrast, applying PBL without adversarial training (see columns 1 \& 3) enhances the standard accuracy of RSVP and preserves or even boosts its robustness. When combining PBL with adversarial training, PBL mitigates the drop in standard accuracy typically induced by adversarial training and sustains robustness enhancements (see columns 2 \& 4), without additional time usage or computational resource consumption.

\section{Conclusion}
In this paper, we thoroughly explore the properties of Robust Source VP (RSVP). We discover that RSVP inherit the robustness of the source model and then we provide an interpretation at visual representation level. Moreover, RSVP also experience suboptimal results in terms of its generalization performance. To address this problem, we introduce a plug-and-play strategy known as Prompt Boundary Loosening (PBL), aiming at reducing the learning difficulty of RSVP by formally relaxing the decision boundary of the source model in conjunction with various label mapping methods. 
Extensive experiments results demonstrate that our findings are universal and the proposed PBL not only maintains the robustness of RSVP but also enhances its generalization ability for various downstream datasets. 

\bibliographystyle{named}
\bibliography{ijcai25}

\newpage
\appendix

\section{Datasets in Detail}
Table \ref{tab:dataset} provides an overview of datasets used in our work. Each dataset is listed with key attributes, reflecting its utility for training and testing models:
\texttt{Flowers102} is dedicated to image classification, this dataset comprises 4,093 training images and 2,463 test images across 102 different flower categories, with images rescaled to a resolution of 128x128 pixels.
\texttt{SVHN} (Street View House Numbers) is utilized for object recognition, it contains 73,257 digits for training and 26,032 for testing, with a class number of 10, representing individual digits from 0 to 9. The images are presented at a resolution of 32x32 pixels.
\texttt{GTSRB} (German Traffic Sign Recognition Benchmark) is another dataset for object recognition, consisting of 39,209 training images and 12,630 test images in 43 classes, representing various traffic signs, also at a resolution of 32x32 pixels.
\texttt{EuroSAT} Used for image classification, this collection features 13,500 training and 8,100 testing satellite images of the Earth, categorized into 10 different classes, with images at a resolution of 128x128 pixels.
\texttt{OxfordPets} is a dataset aimed at object recognition tasks with 2,944 training and 3,669 test images of 37 pet breeds, offered at a resolution of 128x128 pixels.
\texttt{StanfordCars} is designed for image classification and contains 6,509 training images and 8,041 test images of 196 classes of cars, showcased at a resolution of 128x128 pixels.
\texttt{DTD} (Describable Textures Dataset) focused on object recognition, it includes 2,820 training and 1,692 test images across 47 classes, with each image rendered at a resolution of 128x128 pixels.
\texttt{CIFAR100} is a well-known dataset for image classification tasks, featuring 50,000 training and 10,000 test images across 100 classes. The images are provided at a resolution of 32x32 pixels.

\begin{table*}[t!]
  \small
  \begin{center}
  \setlength\tabcolsep{3 pt}
  {\renewcommand{\arraystretch}{0.5}
  \resizebox{\textwidth}{!}{
  \begin{tabular}{llcccccc}
  
  \noalign{\hrule height 0.8pt}
  
  \textbf{Names} & \textbf{Task Descriptions} &\textbf{Train Size} & \textbf{Test Size} & \textbf{Class Number} & \textbf{Rescaled Resolusion} \\
  \noalign{\hrule height 0.5pt} 
  1. \texttt{Flowers102} \cite{nilsback2008automated} & Image Classification & 4093 & 2463 & 102 & 128$\times$128\\
  2. \texttt{SVHN} \cite{netzer2011reading} & Object Recognition & 73257 & 26032 & 10 & 32$\times$32\\
  3. \texttt{GTSRB} \cite{stallkamp2011german} & Object Recognition & 39209 & 12630 & 43 & 32$\times$32 \\
  4. \texttt{EuroSAT} \cite{helber2018introducing,helber2019eurosat}&Image Classification & 13500 & 8100 & 10 & 128$\times$128 \\
  5. \texttt{OxfordPets} \cite{parkhi2012cats} & Object Recognition  & 2944 & 3669 & 37 & 128$\times$128\\
  6. \texttt{StanfordCars} \cite{krause20133d} & Image Classification & 6509 & 8041 & 196 & 128$\times$128 \\

  7. \texttt{DTD} \cite{cimpoi2014describing} & Object Recognition & 2820 & 1692 & 47 & 128$\times$128\\
  8. \texttt{CIFAR100} \cite{krizhevsky2009learning} & Image Classification & 50000 & 10000	& 100 & 32$\times$32 \\
  \noalign{\hrule height 0.8pt}
  
  \end{tabular}}}
  \end{center}
  \caption{Summary of the 8 datasets used in this work. }
  \label{tab:dataset}
  \vspace{-0.5cm}
  \end{table*}

\section{Visualization of RSVP across different setups}
\begin{figure*}[ht!]
    \centering
    \begin{minipage}{\textwidth}
        \centering
        \setcounter{subfigure}{0}
        \begin{subfigure}[b]{0.15\textwidth}
            \centering
            \includegraphics[width=\textwidth]{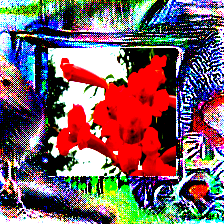}
            \caption{RLM-1}
            \label{fig:sub1a}
        \end{subfigure}
        \hfill
        \begin{subfigure}[b]{0.15\textwidth}
            \centering
            \includegraphics[width=\textwidth]{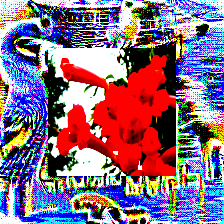}
            \caption{RLM-2}
            \label{fig:sub2a}
        \end{subfigure}
        \hfill
        \begin{subfigure}[b]{0.15\textwidth}
            \centering
            \includegraphics[width=\textwidth]{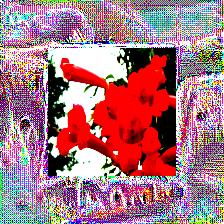}
            \caption{RLM-3}
            \label{fig:sub3a}
        \end{subfigure}
        \hfill
        \begin{subfigure}[b]{0.15\textwidth}
            \centering
            \includegraphics[width=\textwidth]{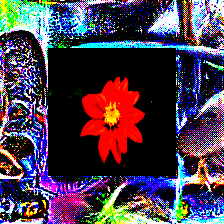}
            \caption{ILM-1}
            \label{fig:sub4a}
        \end{subfigure}
        \hfill
        \begin{subfigure}[b]{0.15\textwidth}
            \centering
            \includegraphics[width=\textwidth]{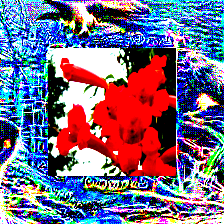}
            \caption{ILM-2}
            \label{fig:sub43a}
        \end{subfigure}
        \hfill
        \begin{subfigure}[b]{0.15\textwidth}
            \centering
            \includegraphics[width=\textwidth]{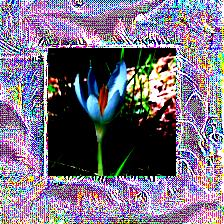}
            \caption{ILM-3}
            \label{fig:sub24a}
        \end{subfigure}
        \captionsetup{labelformat=parens}
    \end{minipage}
    \begin{minipage}{\textwidth}
        \centering
        \setcounter{subfigure}{0}
        \begin{subfigure}[b]{0.15\textwidth}
            \centering
            \includegraphics[width=\textwidth]{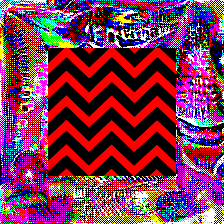}
            \caption{RLM-1}
            \label{fig:sub1a}
        \end{subfigure}
        \hfill
        \begin{subfigure}[b]{0.15\textwidth}
            \centering
            \includegraphics[width=\textwidth]{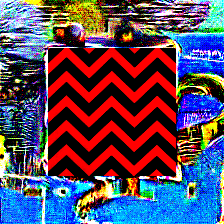}
            \caption{RLM-2}
            \label{fig:sub2a}
        \end{subfigure}
        \hfill
        \begin{subfigure}[b]{0.15\textwidth}
            \centering
            \includegraphics[width=\textwidth]{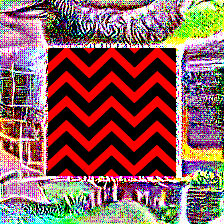}
            \caption{RLM-3}
            \label{fig:sub3a}
        \end{subfigure}
        \hfill
        \begin{subfigure}[b]{0.15\textwidth}
            \centering
            \includegraphics[width=\textwidth]{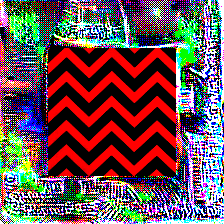}
            \caption{ILM-1}
            \label{fig:sub4a}
        \end{subfigure}
        \hfill
        \begin{subfigure}[b]{0.15\textwidth}
            \centering
            \includegraphics[width=\textwidth]{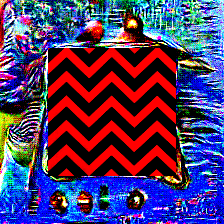}
            \caption{ILM-2}
            \label{fig:sub43a}
        \end{subfigure}
        \hfill
        \begin{subfigure}[b]{0.15\textwidth}
            \centering
            \includegraphics[width=\textwidth]{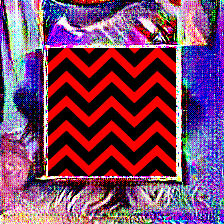}
            \caption{ILM-3}
            \label{fig:sub24a}
        \end{subfigure}
        \captionsetup{labelformat=parens}
    \end{minipage}
    \begin{minipage}{\textwidth}
        \centering
        \setcounter{subfigure}{0}
        \begin{subfigure}[b]{0.15\textwidth}
            \centering
            \includegraphics[width=\textwidth]{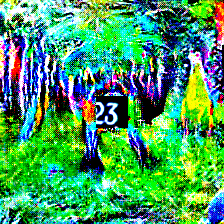}
            \caption{RLM-1}
            \label{fig:sub1a}
        \end{subfigure}
        \hfill
        \begin{subfigure}[b]{0.15\textwidth}
            \centering
            \includegraphics[width=\textwidth]{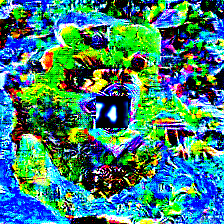}
            \caption{RLM-2}
            \label{fig:sub2a}
        \end{subfigure}
        \hfill
        \begin{subfigure}[b]{0.15\textwidth}
            \centering
            \includegraphics[width=\textwidth]{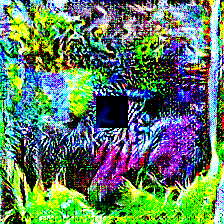}
            \caption{RLM-3}
            \label{fig:sub3a}
        \end{subfigure}
        \hfill
        \begin{subfigure}[b]{0.15\textwidth}
            \centering
            \includegraphics[width=\textwidth]{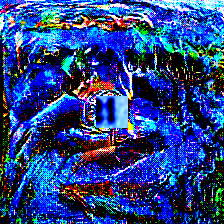}
            \caption{ILM-1}
            \label{fig:sub4a}
        \end{subfigure}
        \hfill
        \begin{subfigure}[b]{0.15\textwidth}
            \centering
            \includegraphics[width=\textwidth]{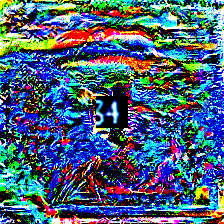}
            \caption{ILM-2}
            \label{fig:sub43a}
        \end{subfigure}
        \hfill
        \begin{subfigure}[b]{0.15\textwidth}
            \centering
            \includegraphics[width=\textwidth]{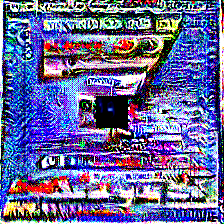}
            \caption{ILM-3}
            \label{fig:sub24a}
        \end{subfigure}
        \captionsetup{labelformat=parens}
    \end{minipage}
    \begin{minipage}{\textwidth}
        \centering
        \setcounter{subfigure}{0}
        \begin{subfigure}[b]{0.15\textwidth}
            \centering
            \includegraphics[width=\textwidth]{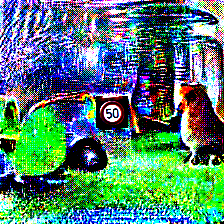}
            \caption{RLM-1}
            \label{fig:sub1a}
        \end{subfigure}
        \hfill
        \begin{subfigure}[b]{0.15\textwidth}
            \centering
            \includegraphics[width=\textwidth]{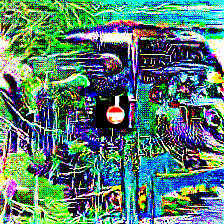}
            \caption{RLM-2}
            \label{fig:sub2a}
        \end{subfigure}
        \hfill
        \begin{subfigure}[b]{0.15\textwidth}
            \centering
            \includegraphics[width=\textwidth]{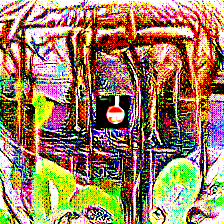}
            \caption{RLM-3}
            \label{fig:sub3a}
        \end{subfigure}
        \hfill
        \begin{subfigure}[b]{0.15\textwidth}
            \centering
            \includegraphics[width=\textwidth]{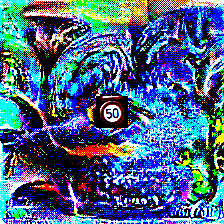}
            \caption{ILM-1}
            \label{fig:sub4a}
        \end{subfigure}
        \hfill
        \begin{subfigure}[b]{0.15\textwidth}
            \centering
            \includegraphics[width=\textwidth]{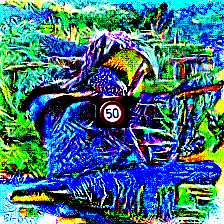}
            \caption{ILM-2}
            \label{fig:sub43a}
        \end{subfigure}
        \hfill
        \begin{subfigure}[b]{0.15\textwidth}
            \centering
            \includegraphics[width=\textwidth]{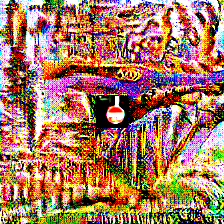}
            \caption{ILM-3}
            \label{fig:sub24a}
        \end{subfigure}
        \captionsetup{labelformat=parens}
    \end{minipage}
    \caption{Visualization of RSVP obtained when different robust models are used as source models. Each row from left to right: the first three are the results of three different robust models under RLM, and the last three are the results of three different robust models under ILM. Four lines represent the result of: Flowers102, DTD, SVHN, GTSRB dataset from top to bottom respectively.}
    \label{fig:overall_1}
\end{figure*}

\begin{figure*}[ht!]
    \centering
    \begin{minipage}{\textwidth}
        \centering
        \setcounter{subfigure}{0}
        \begin{subfigure}[b]{0.15\textwidth}
            \centering
            \includegraphics[width=\textwidth]{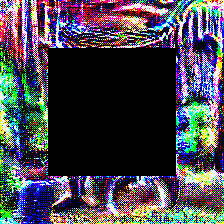}
            \caption{RLM-1}
            \label{fig:sub1a}
        \end{subfigure}
        \hfill
        \begin{subfigure}[b]{0.15\textwidth}
            \centering
            \includegraphics[width=\textwidth]{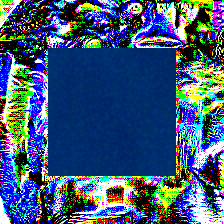}
            \caption{RLM-2}
            \label{fig:sub2a}
        \end{subfigure}
        \hfill
        \begin{subfigure}[b]{0.15\textwidth}
            \centering
            \includegraphics[width=\textwidth]{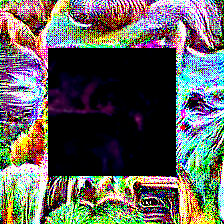}
            \caption{RLM-3}
            \label{fig:sub3a}
        \end{subfigure}
        \hfill
        \begin{subfigure}[b]{0.15\textwidth}
            \centering
            \includegraphics[width=\textwidth]{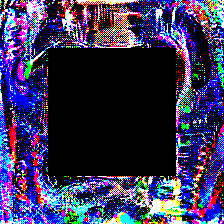}
            \caption{ILM-1}
            \label{fig:sub4a}
        \end{subfigure}
        \hfill
        \begin{subfigure}[b]{0.15\textwidth}
            \centering
            \includegraphics[width=\textwidth]{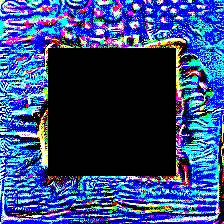}
            \caption{ILM-2}
            \label{fig:sub43a}
        \end{subfigure}
        \hfill
        \begin{subfigure}[b]{0.15\textwidth}
            \centering
            \includegraphics[width=\textwidth]{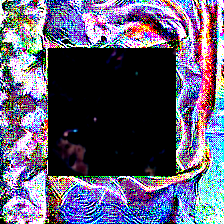}
            \caption{ILM-3}
            \label{fig:sub24a}
        \end{subfigure}
        \captionsetup{labelformat=parens}
    \end{minipage}
    \begin{minipage}{\textwidth}
        \centering
        \setcounter{subfigure}{0}
        \begin{subfigure}[b]{0.15\textwidth}
            \centering
            \includegraphics[width=\textwidth]{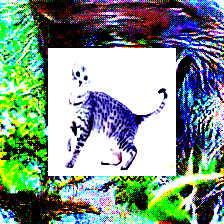}
            \caption{RLM-1}
            \label{fig:sub1a}
        \end{subfigure}
        \hfill
        \begin{subfigure}[b]{0.15\textwidth}
            \centering
            \includegraphics[width=\textwidth]{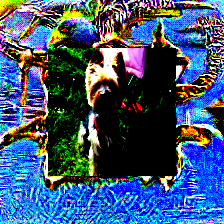}
            \caption{RLM-2}
            \label{fig:sub2a}
        \end{subfigure}
        \hfill
        \begin{subfigure}[b]{0.15\textwidth}
            \centering
            \includegraphics[width=\textwidth]{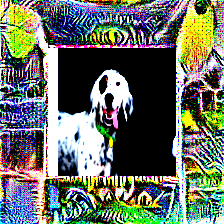}
            \caption{RLM-3}
            \label{fig:sub3a}
        \end{subfigure}
        \hfill
        \begin{subfigure}[b]{0.15\textwidth}
            \centering
            \includegraphics[width=\textwidth]{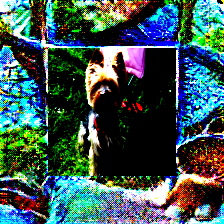}
            \caption{ILM-1}
            \label{fig:sub4a}
        \end{subfigure}
        \hfill
        \begin{subfigure}[b]{0.15\textwidth}
            \centering
            \includegraphics[width=\textwidth]{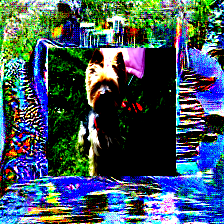}
            \caption{ILM-2}
            \label{fig:sub43a}
        \end{subfigure}
        \hfill
        \begin{subfigure}[b]{0.15\textwidth}
            \centering
            \includegraphics[width=\textwidth]{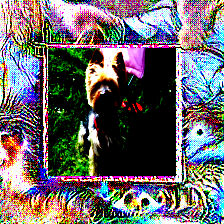}
            \caption{ILM-3}
            \label{fig:sub24a}
        \end{subfigure}
        \captionsetup{labelformat=parens}
    \end{minipage}
    \begin{minipage}{\textwidth}
        \centering
        \setcounter{subfigure}{0}
        \begin{subfigure}[b]{0.15\textwidth}
            \centering
            \includegraphics[width=\textwidth]{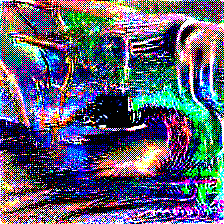}
            \caption{RLM-1}
            \label{fig:sub1a}
        \end{subfigure}
        \hfill
        \begin{subfigure}[b]{0.15\textwidth}
            \centering
            \includegraphics[width=\textwidth]{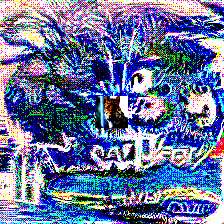}
            \caption{RLM-2}
            \label{fig:sub2a}
        \end{subfigure}
        \hfill
        \begin{subfigure}[b]{0.15\textwidth}
            \centering
            \includegraphics[width=\textwidth]{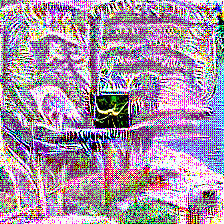}
            \caption{RLM-3}
            \label{fig:sub3a}
        \end{subfigure}
        \hfill
        \begin{subfigure}[b]{0.15\textwidth}
            \centering
            \includegraphics[width=\textwidth]{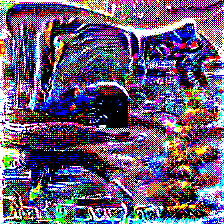}
            \caption{ILM-1}
            \label{fig:sub4a}
        \end{subfigure}
        \hfill
        \begin{subfigure}[b]{0.15\textwidth}
            \centering
            \includegraphics[width=\textwidth]{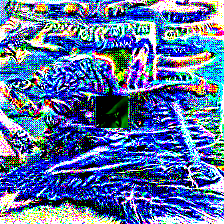}
            \caption{ILM-2}
            \label{fig:sub43a}
        \end{subfigure}
        \hfill
        \begin{subfigure}[b]{0.15\textwidth}
            \centering
            \includegraphics[width=\textwidth]{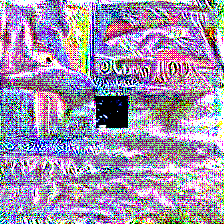}
            \caption{ILM-3}
            \label{fig:sub24a}
        \end{subfigure}
        \captionsetup{labelformat=parens}
    \end{minipage}
    \begin{minipage}{\textwidth}
        \centering
        \setcounter{subfigure}{0}
        \begin{subfigure}[b]{0.15\textwidth}
            \centering
            \includegraphics[width=\textwidth]{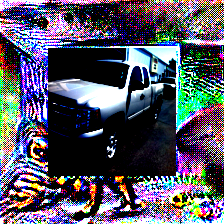}
            \caption{RLM-1}
            \label{fig:sub1a}
        \end{subfigure}
        \hfill
        \begin{subfigure}[b]{0.15\textwidth}
            \centering
            \includegraphics[width=\textwidth]{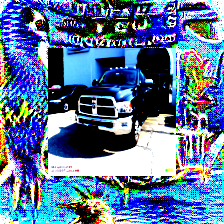}
            \caption{RLM-2}
            \label{fig:sub2a}
        \end{subfigure}
        \hfill
        \begin{subfigure}[b]{0.15\textwidth}
            \centering
            \includegraphics[width=\textwidth]{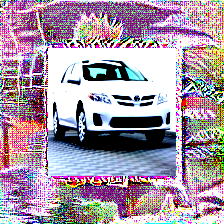}
            \caption{RLM-3}
            \label{fig:sub3a}
        \end{subfigure}
        \hfill
        \begin{subfigure}[b]{0.15\textwidth}
            \centering
            \includegraphics[width=\textwidth]{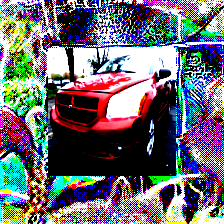}
            \caption{ILM-1}
            \label{fig:sub4a}
        \end{subfigure}
        \hfill
        \begin{subfigure}[b]{0.15\textwidth}
            \centering
            \includegraphics[width=\textwidth]{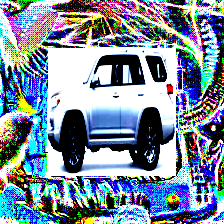}
            \caption{ILM-2}
            \label{fig:sub43a}
        \end{subfigure}
        \hfill
        \begin{subfigure}[b]{0.15\textwidth}
            \centering
            \includegraphics[width=\textwidth]{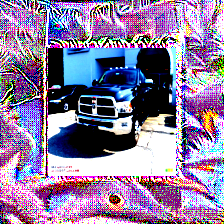}
            \caption{ILM-3}
            \label{fig:sub24a}
        \end{subfigure}
        \captionsetup{labelformat=parens}
    \end{minipage}
    \caption{Visualization of RSVP obtained when different robust models are used as source models. Each row from left to right: the first three are the results of three different robust models under RLM, and the last three are the results of three different robust models under ILM. Four lines represent the result of: EuroSAT, OxfordPets, CIFAR100, StanfordCars dataset from top to bottom respectively.}
    \label{fig:overall_2}
\end{figure*}

Figure \ref{fig:overall_1} and Figure \ref{fig:overall_2} show the visualization results of RSVP under different settings. In this experiment, three different robust source models are used: Same as in the main manuscript, we select one standard model and three robust models trained with ImageNet \cite{deng2009imagenet} under the $l_{\infty}$-norm \cite{salman2020adversarially,robustness,wong2020fast}.

Figure \ref{fig:overall_1} presents a comparative visualization of RSVP outcomes when applying various robust models as source models. Displayed in rows, the first trio of images from left to right depict the results obtained from three distinct robust models using the Random Label Model (RLM) approach. The subsequent trio showcases outcomes from the Iterative Label Model (ILM) strategy. Each row corresponds to a specific dataset, with the sequence from top to bottom representing the results for the Flowers102, DTD, SVHN, and GTSRB datasets, respectively. The visualizations across all datasets demonstrate variations influenced by the choice of robust source model and the label mapping strategies employed. However, all outcomes reveal the distinct characteristics of RSVP as compared to SSVP: they are more in accordance with human perception. Besides, for the same robust source model, the visualization results tend to share a similar pattern, such as the third row and the fourth row of (d), they all take \cite{salman2020adversarially} as the source model and use the same label mapping strategy (ILM). 
The visualizations yield discernible elements that mirror real-world objects. For example, in the first row of (b), there appears to be an avian figure on the left side of the RSVP sequence. Similarly, in the third row of image (f), a distinct geometric shape, reminiscent of the letter 'Z', is clearly identifiable.
Figure \ref{fig:overall_2} presents a comparative visualization of the rest four datasets. Same as Figure \ref{fig:overall_1}, the first trio of images from left to right depict the results obtained from three distinct robust models using the Random Label Model (RLM) approach. The subsequent trio showcases outcomes from the Iterative Label Model (ILM) strategy. Each row corresponds to a specific dataset, with the sequence from top to bottom representing the results for the EuroSAT, Oxfordpets, CIFAR100, and StanfordCars datasets, respectively.The insights drawn here echo those presented in Figure \ref{fig:overall_1}.
\begin{figure*}[ht!]
    \centering
    \begin{minipage}{\textwidth}
        \centering
        \setcounter{subfigure}{0}
        \begin{subfigure}[b]{0.1\textwidth}
            \centering
            \includegraphics[width=\textwidth]{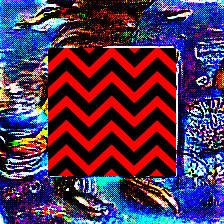}
            \caption{RLM-5}
            \label{fig:sub1a}
        \end{subfigure}
        \hfill
        \begin{subfigure}[b]{0.1\textwidth}
            \centering
            \includegraphics[width=\textwidth]{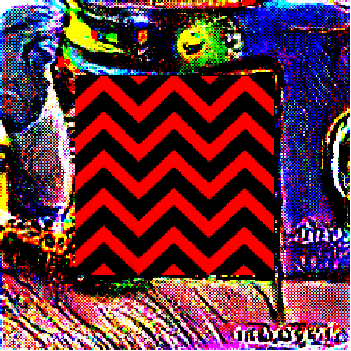}
            \caption{RLM-10}
            \label{fig:sub2a}
        \end{subfigure}
        \hfill
        \begin{subfigure}[b]{0.1\textwidth}
            \centering
            \includegraphics[width=\textwidth]{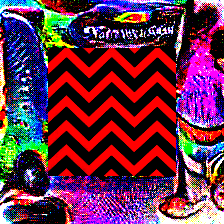}
            \caption{RLM-15}
            \label{fig:sub3a}
        \end{subfigure}
        \hfill
        \begin{subfigure}[b]{0.1\textwidth}
            \centering
            \includegraphics[width=\textwidth]{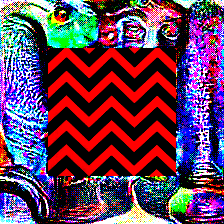}
            \caption{RLM-20}
            \label{fig:sub4a}
        \end{subfigure}
        \hfill
        \begin{subfigure}[b]{0.1\textwidth}
            \centering
            \includegraphics[width=\textwidth]{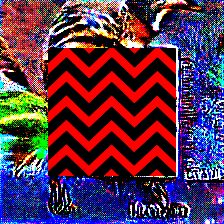}
            \caption{ILM-5}
            \label{fig:sub43a}
        \end{subfigure}
        \hfill
        \begin{subfigure}[b]{0.1\textwidth}
            \centering
            \includegraphics[width=\textwidth]{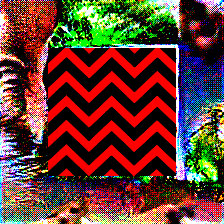}
            \caption{ILM-10}
            \label{fig:sub24a}
        \end{subfigure}
        \hfill
        \begin{subfigure}[b]{0.1\textwidth}
            \centering
            \includegraphics[width=\textwidth]{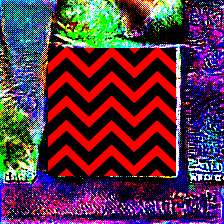}
            \caption{ILM-15}
            \label{fig:sub24a}
        \end{subfigure}
        \hfill
        \begin{subfigure}[b]{0.1\textwidth}
            \centering
            \includegraphics[width=\textwidth]{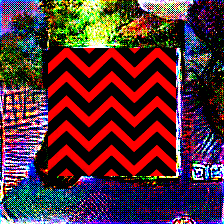}
            \caption{ILM-20}
            \label{fig:sub24a}
        \end{subfigure}
        \captionsetup{labelformat=parens}
    \end{minipage}
    \begin{minipage}{\textwidth}
        \centering
        \setcounter{subfigure}{0}
        \begin{subfigure}[b]{0.1\textwidth}
            \centering
            \includegraphics[width=\textwidth]{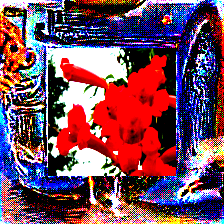}
            \caption{RLM-3}
            \label{fig:sub1a}
        \end{subfigure}
        \hfill
        \begin{subfigure}[b]{0.1\textwidth}
            \centering
            \includegraphics[width=\textwidth]{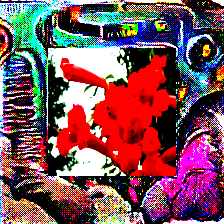}
            \caption{RLM-5}
            \label{fig:sub2a}
        \end{subfigure}
        \hfill
        \begin{subfigure}[b]{0.1\textwidth}
            \centering
            \includegraphics[width=\textwidth]{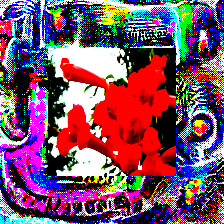}
            \caption{RLM-7}
            \label{fig:sub3a}
        \end{subfigure}
        \hfill
        \begin{subfigure}[b]{0.1\textwidth}
            \centering
            \includegraphics[width=\textwidth]{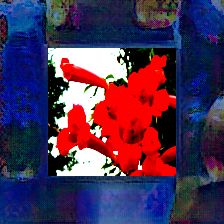}
            \caption{RLM-9}
            \label{fig:sub4a}
        \end{subfigure}
        \hfill
        \begin{subfigure}[b]{0.1\textwidth}
            \centering
            \includegraphics[width=\textwidth]{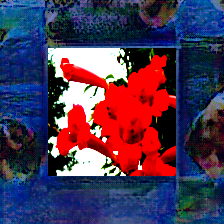}
            \caption{ILM-3}
            \label{fig:sub43a}
        \end{subfigure}
        \hfill
        \begin{subfigure}[b]{0.1\textwidth}
            \centering
            \includegraphics[width=\textwidth]{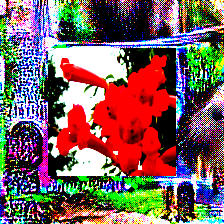}
            \caption{ILM-5}
            \label{fig:sub24a}
        \end{subfigure}
        \hfill
        \begin{subfigure}[b]{0.1\textwidth}
            \centering
            \includegraphics[width=\textwidth]{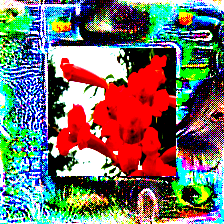}
            \caption{ILM-7}
            \label{fig:sub24a}
        \end{subfigure}
        \hfill
        \begin{subfigure}[b]{0.1\textwidth}
            \centering
            \includegraphics[width=\textwidth]{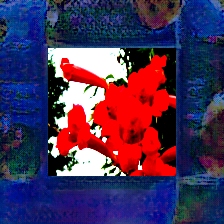}
            \caption{ILM-9}
            \label{fig:sub24a}
        \end{subfigure}
        \captionsetup{labelformat=parens}
    \end{minipage}
    \begin{minipage}{\textwidth}
        \centering
        \setcounter{subfigure}{0}
        \begin{subfigure}[b]{0.1\textwidth}
            \centering
            \includegraphics[width=\textwidth]{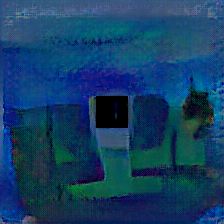}
            \caption{RLM-10}
            \label{fig:sub1a}
        \end{subfigure}
        \hfill
        \begin{subfigure}[b]{0.1\textwidth}
            \centering
            \includegraphics[width=\textwidth]{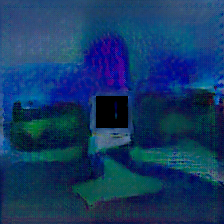}
            \caption{RLM-20}
            \label{fig:sub2a}
        \end{subfigure}
        \hfill
        \begin{subfigure}[b]{0.1\textwidth}
            \centering
            \includegraphics[width=\textwidth]{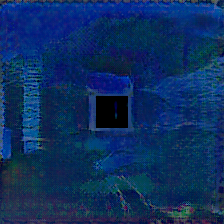}
            \caption{RLM-30}
            \label{fig:sub3a}
        \end{subfigure}
        \hfill
        \begin{subfigure}[b]{0.1\textwidth}
            \centering
            \includegraphics[width=\textwidth]{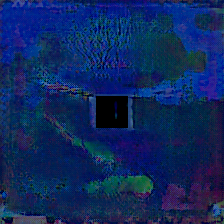}
            \caption{RLM-40}
            \label{fig:sub4a}
        \end{subfigure}
        \hfill
        \begin{subfigure}[b]{0.1\textwidth}
            \centering
            \includegraphics[width=\textwidth]{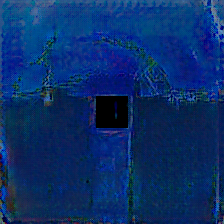}
            \caption{ILM-10}
            \label{fig:sub43a}
        \end{subfigure}
        \hfill
        \begin{subfigure}[b]{0.1\textwidth}
            \centering
            \includegraphics[width=\textwidth]{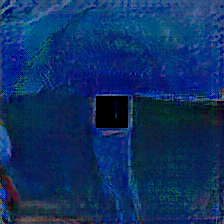}
            \caption{ILM-20}
            \label{fig:sub24a}
        \end{subfigure}
        \hfill
        \begin{subfigure}[b]{0.1\textwidth}
            \centering
            \includegraphics[width=\textwidth]{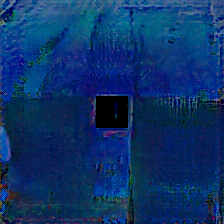}
            \caption{ILM-30}
            \label{fig:sub24a}
        \end{subfigure}
        \hfill
        \begin{subfigure}[b]{0.1\textwidth}
            \centering
            \includegraphics[width=\textwidth]{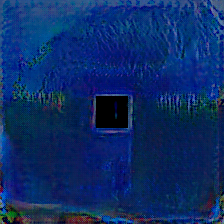}
            \caption{ILM-40}
            \label{fig:sub24a}
        \end{subfigure}
        \captionsetup{labelformat=parens}
    \end{minipage}
    \begin{minipage}{\textwidth}
        \centering
        \setcounter{subfigure}{0}
        \begin{subfigure}[b]{0.1\textwidth}
            \centering
            \includegraphics[width=\textwidth]{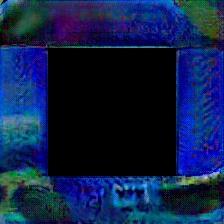}
            \caption{RLM-10}
            \label{fig:sub1a}
        \end{subfigure}
        \hfill
        \begin{subfigure}[b]{0.1\textwidth}
            \centering
            \includegraphics[width=\textwidth]{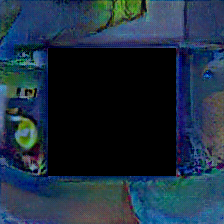}
            \caption{RLM-20}
            \label{fig:sub2a}
        \end{subfigure}
        \hfill
        \begin{subfigure}[b]{0.1\textwidth}
            \centering
            \includegraphics[width=\textwidth]{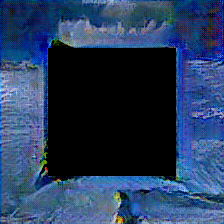}
            \caption{RLM-30}
            \label{fig:sub3a}
        \end{subfigure}
        \hfill
        \begin{subfigure}[b]{0.1\textwidth}
            \centering
            \includegraphics[width=\textwidth]{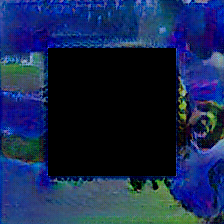}
            \caption{RLM-40}
            \label{fig:sub4a}
        \end{subfigure}
        \hfill
        \begin{subfigure}[b]{0.1\textwidth}
            \centering
            \includegraphics[width=\textwidth]{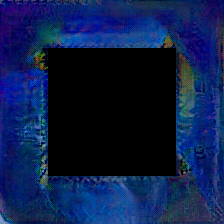}
            \caption{ILM-10}
            \label{fig:sub43a}
        \end{subfigure}
        \hfill
        \begin{subfigure}[b]{0.1\textwidth}
            \centering
            \includegraphics[width=\textwidth]{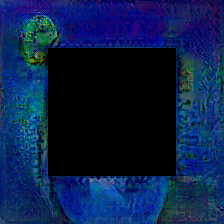}
            \caption{ILM-20}
            \label{fig:sub24a}
        \end{subfigure}
        \hfill
        \begin{subfigure}[b]{0.1\textwidth}
            \centering
            \includegraphics[width=\textwidth]{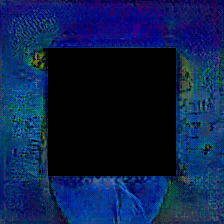}
            \caption{ILM-30}
            \label{fig:sub24a}
        \end{subfigure}
        \hfill
        \begin{subfigure}[b]{0.1\textwidth}
            \centering
            \includegraphics[width=\textwidth]{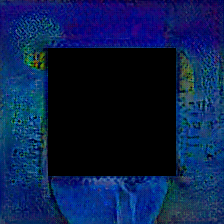}
            \caption{ILM-40}
            \label{fig:sub24a}
        \end{subfigure}
        \captionsetup{labelformat=parens}
    \end{minipage}
    \begin{minipage}{\textwidth}
        \centering
        \setcounter{subfigure}{0}
        \begin{subfigure}[b]{0.1\textwidth}
            \centering
            \includegraphics[width=\textwidth]{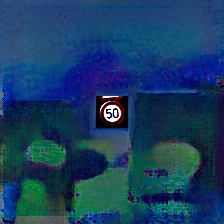}
            \caption{RLM-5}
            \label{fig:sub1a}
        \end{subfigure}
        \hfill
        \begin{subfigure}[b]{0.1\textwidth}
            \centering
            \includegraphics[width=\textwidth]{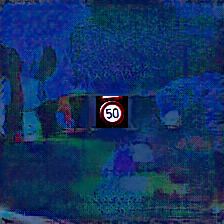}
            \caption{RLM-10}
            \label{fig:sub2a}
        \end{subfigure}
        \hfill
        \begin{subfigure}[b]{0.1\textwidth}
            \centering
            \includegraphics[width=\textwidth]{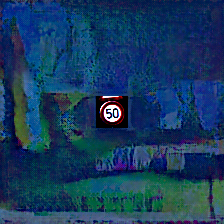}
            \caption{RLM-15}
            \label{fig:sub3a}
        \end{subfigure}
        \hfill
        \begin{subfigure}[b]{0.1\textwidth}
            \centering
            \includegraphics[width=\textwidth]{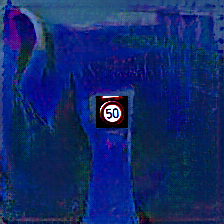}
            \caption{RLM-20}
            \label{fig:sub4a}
        \end{subfigure}
        \hfill
        \begin{subfigure}[b]{0.1\textwidth}
            \centering
            \includegraphics[width=\textwidth]{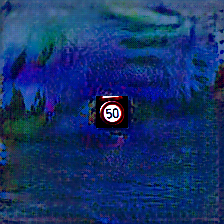}
            \caption{ILM-5}
            \label{fig:sub43a}
        \end{subfigure}
        \hfill
        \begin{subfigure}[b]{0.1\textwidth}
            \centering
            \includegraphics[width=\textwidth]{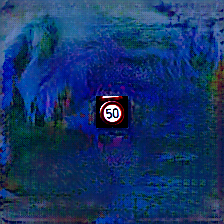}
            \caption{ILM-10}
            \label{fig:sub24a}
        \end{subfigure}
        \hfill
        \begin{subfigure}[b]{0.1\textwidth}
            \centering
            \includegraphics[width=\textwidth]{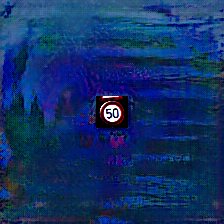}
            \caption{ILM-15}
            \label{fig:sub24a}
        \end{subfigure}
        \hfill
        \begin{subfigure}[b]{0.1\textwidth}
            \centering
            \includegraphics[width=\textwidth]{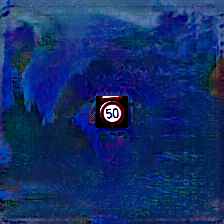}
            \caption{ILM-20}
            \label{fig:sub24a}
        \end{subfigure}
        \captionsetup{labelformat=parens}
    \end{minipage}
    \caption{Visualization of RSVP obtained when utilizing the proposed PBL method. Each row from left to right: the first four are the results under four different temporeture $\mathcal{T}$ when using RLM as label mapping method, and the last four are the results of same four temporeture $\mathcal{T}$ under ILM. Five lines represent the result of: DTD, Flowers102, SVHN, EuroSAT and GTSRB dataset from top to bottom respectively. For DTD, $\mathcal{T}$ is set to be: 5, 10, 15 and 20; for Flowers102, $\mathcal{T}$ is set to be: 3, 5, 7, 9; for SVHN, $\mathcal{T}$ is set to be: 10, 20, 30, 40; for EuroSAT, $\mathcal{T}$ is set to be: 10, 20, 30, 40; for GTSRB, $\mathcal{T}$ is set to be: 5, 10, 15, 20.}
    \label{fig:pbl_vis}
\end{figure*}

\section{Visualization of RSVP when utilizing PBL}
The visualizations displayed in Figure \ref{fig:pbl_vis} are generated using the proposed PBL method. For each dataset, the first four RSVP in a row illustrate the results at varying temperature settings (\(\mathcal{T}\)) using the RLM as label mapping strategy, while the subsequent four RSVP show the outcomes for the identical temperatures under the ILM method. The sequences of images from top to bottom correspond to the DTD, Flowers102, SVHN, EuroSAT, and GTSRB datasets, respectively. Specifically, for DTD, the temperatures are set at 5, 10, 15, and 20. For Flowers102, they are 3, 5, 7, and 9. SVHN and EuroSAT both have temperature settings of 10, 20, 30, and 40. Lastly, for GTSRB, the temperatures are set at 5, 10, 15, and 20. It can be observed that the RSVP still has a clear human-aligned visualization after utilizing PBL, but the pattern is different from that without PBL: 
These visualizations tend to favor a darker color scheme and when temperature is larger, the RSVP will exhibit certain consistency, showing minimal variation across different temperature settings.
\end{document}